\documentclass{article}

\PassOptionsToPackage{numbers,compress}{natbib}
 \usepackage[preprint]{neurips_2026}

\usepackage[utf8]{inputenc} 
\usepackage[T1]{fontenc}    
\usepackage{hyperref}       
\usepackage{url}            
\usepackage{booktabs}       
\usepackage{amsmath}
\usepackage{amsfonts}       
\usepackage{nicefrac}       
\usepackage{microtype}      
\usepackage{xcolor}         
\usepackage[pdftex]{graphicx}
\usepackage{subcaption}
\usepackage{cleveref}
\usepackage{algorithm}
\usepackage{algpseudocode}
\usepackage{wrapfig}
\usepackage{tikz, forest}
\usetikzlibrary{bayesnet}
\usepackage{pifont}
\newcommand{\E}{\mathbb{E}}

\newcommand{\KL}{D_{\mathrm{KL}}}

\definecolor{pastelblue}{RGB}{90, 140, 200}   
\definecolor{pastelgreen}{RGB}{90, 180, 90}   
\definecolor{pastelred}{RGB}{200, 100, 100}   
\definecolor{darkgreen}{HTML}{2CA02C}
\definecolor{pastelyellow}{RGB}{139, 128, 0}   
\definecolor{pastelpurple}{RGB}{195, 177, 225}   

\title{
JEDI: Joint Embedding Diffusion World Model for Online Model-Based Reinforcement Learning}

\author{%
  Jing Yu Lim \\
  \And
  Rushi Shah\\
  \And
  Zarif Ikram\\
  \And
   Samson Yu\\
   \And
   Haozhe Ma\\
   \And
   Tze-Yun Leong\\
   \And
   Dianbo Liu\\\\
   National University of Singapore\\
   \texttt{\{jing.yu, shah.15, leongty, dianbo\}@nus.edu.sg}, 
   \texttt{zikram@usc.edu},\\ 
   \texttt{\{samson.yu, haozhe.ma\}@u.nus.edu.sg}\\
}

\begin{document}

\maketitle

\begin{abstract}
Diffusion world models have recently become competitive for online model-based reinforcement learning, but current approaches expose a tension: pixel diffusion is effective but computationally expensive while the latest latent diffusion approach improves efficiency yet performs subpar. The latter also relies on separately trained latents rather than the end-to-end world-model objectives that have driven much of modern MBRL progress. In particular, JEPA-style predictive representation learning has emerged as an especially promising direction for world modeling and MBRL. Concurrently, diffusion-style objectives have gained traction across multiple domains, with iterative refinement as a promising approach for multimodal and stochastic targets. Taken together, these trends motivate Joint Embedding DIffusion (JEDI), the first online end-to-end latent diffusion world model. JEDI learns its latent space directly from the diffusion denoising loss with a JEPA framework, using denoising to learn and predict future latents rather than relying on reconstruction and pretrained models. We provide a theoretical motivation showing that conventional JEPA objectives induce a predictive information bottleneck, and that conditional diffusion denoising admits a closely related predictive-compression decomposition. Empirically, JEDI is competitive on Atari100k and outperforms the baseline with seperately trained latents where directly comparable. Relative to the pixel diffusion baseline, JEDI uses 43\% less VRAM, over 3$\times$ faster world-model sampling, and 2.5$\times$ faster training. JEDI also exhibits a markedly different task-level performance profile from the pixel baseline, suggesting that end-to-end predictive latents change more than compute alone.
\end{abstract}

\section{Introduction}
\label{sec:intro}
Model-based reinforcement learning (MBRL) uses a learned world model \citep{sutton1991DYNA,ha2018World} to predict future outcomes and improve sample efficiency for training an online RL agent \citep{Hafner2020Dreamer}. Diffusion models have recently emerged as a powerful backbone for world modeling, especially in visually complex domains \citep{liu2024sora,parkerholder2024genie2,blattmann2023StableVideoDiff}. However, the strongest online diffusion world models for MBRL, DIAMOND, still operate in pixel space, where denoising is expensive in memory, sampling time, and training time \citep{alonso2024diamond}. A natural next step is therefore to move diffusion into latent space.

Recent latent diffusion approaches such as Horizon Imagination (HI) \citep{cohen2026horizonImagination} show that this can improve efficiency, but they learn their latents without end-to-end sequential prediction and do not yet match the performance of DIAMOND \citep{alonso2024diamond}. This leaves a clear opportunity: if much of the past progress in MBRL has come from end-to-end representation learning with world-model objectives, then latent diffusion may benefit from the same principle \citep{hafner2019Planet,Hafner2020Dreamer,hafner2020Dreamerv2,hafner2023Dreamerv3,zhang2023storm,robine2023TWM}. 

One especially promising way to instantiate the principle of end-to-end predictive representation learning is through Joint Embedding Predictive Architecture (JEPA). 
The goal is to learn representations that capture the predictable, abstract structure of the world while avoiding the need to reconstruct every low-level or intrinsically unpredictable detail \citep{PMAX,grill2020BYOL,lecun2022path,assran2023I-JEPA,bardes2024vjepa}. That inductive bias is especially appealing for MBRL, where the latent should retain information useful for future prediction, rewards, and control, while discarding nuisance variation that is irrelevant for decision making \citep{hansen2022TDMPC,hansen2023tdmpc2,assran2025VJEPA2, wang2026temporalstraightening,maes2026leworldmodel}.

In parallel, there has been a general shift toward diffusion-style objectives rather than one-shot prediction across several domains. These include policy learning \citep{chi2023DiffusionPolicy,wang2022DiffPolicyExpressive, frans2025diffusionGuidance}, language modeling \citep{lou2023discreteDiffLMSEDD, prabhudesai2507diffusionBeatLM, li2022diffusion-lm}, and even classification \citep{belhasin2025advancing}. These works suggest that diffusion is especially useful for multimodal, stochastic, or sequential targets, where iterative denoising can support more expressive prediction and greater robustness to uncertainty. In the context of MBRL, this makes diffusion a relevant objective to examine for end-to-end latent world models.

\begin{figure*}[t]
    \centering
    \begin{subfigure}{0.35\textwidth}
        \centering
        \includegraphics[width=\textwidth]{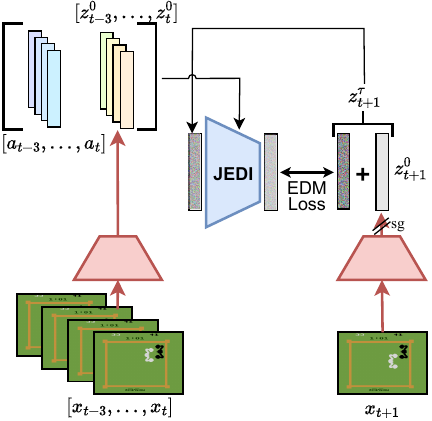}
        \caption{JEDI training}
        \label{fig:JEDI_train}
    \end{subfigure}
    \hspace{10pt}
    \vrule width 0.5pt
    \hspace{10pt}
    \begin{subfigure}{0.35\textwidth}
        \centering
        \includegraphics[width=\textwidth]{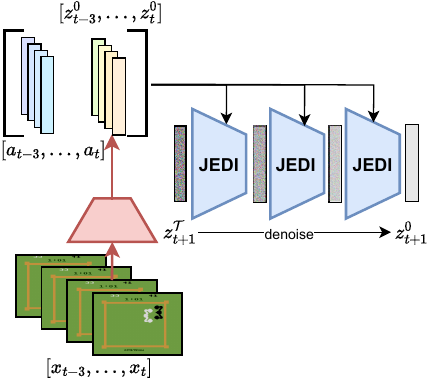}
        \caption{JEDI inference}
        \label{fig:JEDI_inf}
    \end{subfigure}
    \caption{Joint Embedding DIffusion (JEDI) world model. During training, observations are encoded into clean latents, the next latent is noised, and the denoiser predicts the direction back toward the clean future latent while gradients update the encoder end to end. During inference, the model conditions on past latents and actions, starts from noise, and denoises to the next latent for imagination and policy learning.}
    \label{fig:jedi_training_inference}
\end{figure*}

Taken together, these observations raise a natural question: can a diffusion world model learn its latent representation end to end from the denoising objective with JEPA? We answer yes. We introduce Joint Embedding DIffusion (JEDI), a latent diffusion world model that trains its encoder end-to-end through conditional denoising, together with reward and end prediction (\cref{fig:jedi_training_inference}). To our knowledge, JEDI is the first online end-to-end latent diffusion world model built on a JEPA framework.

We show that this representation-learning strategy is theoretically motivated (\cref{sec:theory}): latent conditional diffusion admits a variational information-bottleneck decomposition \citep{alemi2016deepVarIB, tishby2000information, shwartz2017opening}, helping explain why latent conditional denoising can support end-to-end latent learning in a world model. 

Empirically, JEDI delivers strong results on full Atari100k \citep{kaiser2019Atari100kSimPLE} and in supporting Craftium \citep{malagon2024craftium} experiments (\cref{fig:main_result,fig:Atari_JEDI_HI_DIAMOND,fig:craftium_JEDI_HI}) while dramatically improving efficiency over DIAMOND (\cref{fig:compute_and_runtime_comparison,tab:wall_clock_runtime}), using 43\% less VRAM, more than 3$\times$ faster world-model sampling, and more than 2.5$\times$ faster training.

More broadly, these results suggest that \emph{predictive} latent diffusion is a promising alternative to latents trained separately with reconstruction and perceptual losses for online MBRL. While DIAMOND remains a strong baseline, our results suggest that end-to-end predictive latents may offer a more attractive path forward when both efficiency and performance matter. Finally, we also observe a markedly different performance profile across games relative to DIAMOND (\cref{fig:task_performance_profile,fig:hns_performance_profile}), suggesting that end-to-end predictive latents change more than compute alone (\cref{sec:discussion}).
\\
\section{Theoretical Motivation: Information Bottleneck in JEPA}
\label{sec:theory}
\begin{wrapfigure}{r}{0.15\textwidth}
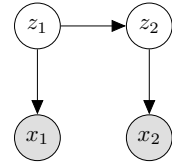

    \vspace{-10pt}
    \centering
    \resizebox{\linewidth}{!}{%
    \tikz{%
    \node[obs]                      (x1) {$x_1$} ; %
    \node[latent, above=of x1]      (z1)  {$z_1$} ;%
    \node[latent, right=of z1]       (z2) {$z_2$} ; %
    \node[obs, right=of x1]       (x2) {$x_2$} ; %

    \edge {z1} {x1} ; %
    \edge {z1} {z2} ; %
    \edge {z2} {x2} ; %
    }%
    }
    \caption{JEPA PGM}
    \label{fig:jepa_pgm}

\end{wrapfigure}
We show that JEPA can be given a variational information-bottleneck interpretation \citep{shwartz2017opening,tishby2000information, alemi2016deepVarIB} complementary to prior formulations for self-supervised representation learning \citep{2023multiviewJepa, shwartz2024compress}. With the Probabilistic Graphical Model (PGM) in \cref{fig:jepa_pgm}, the bottleneck structure emerges directly from the JEPA predictive objective:
\begin{equation}
\mathcal L_{\mathrm{JEPA}} := \E_{p(x_1,x_2)q_\phi(z_1\mid x_1)}\!\left[\KL\bigl(q_\phi(z_2\mid x_2)\,\|\,p_\theta(z_2\mid z_1)\bigr)\right].
\label{eq:jepa_loss}
\end{equation}
Here, $x_1$ and $x_2$ can denote sequential observations in an MDP, with the action omitted for clarity, or two views in a multiview SSL setting. In practice, this KL reduces to cross-entropy for categorical targets and to mean-squared error for fixed-variance Gaussian targets, corresponding to existing works \citep{caron2021DINO, assran2023I-JEPA}. This KL term is also common in canonical end-to-end MBRL works \citep{Hafner2020Dreamer,hafner2020Dreamerv2, hafner2023Dreamerv3, hansen2022TDMPC, hansen2023tdmpc2}. Appendix \ref{sec:proof} makes this precise and shows that the JEPA objective can be written as
\begin{equation}
-\mathcal L_{\mathrm{JEPA}} = I(X_1;X_2) - \widehat I(X_1;Z_1) - \widehat I(X_2;Z_2) + \mathcal R_1 - \mathcal G,
\label{eq:main_jepa_ib}
\end{equation}
where $\widehat I(X_i;Z_i)$ are variational mutual-information estimates, $\mathcal R_1$ is the bottleneck regularizer, and $\mathcal G$ is the posterior approximation gap. Using the data processing inequality yields the bound:
\begin{equation}
-\mathcal L_{\mathrm{JEPA}} \le I(Z_1;Z_2) - \widehat I(X_1;Z_1) - \widehat I(X_2;Z_2) + \mathcal R_1 - \mathcal G.
\label{eq:main_jepa_ib_bound}
\end{equation}
The decomposition makes an information-bottleneck structure explicit. More precisely, it shows that the predictive JEPA loss appears inside a variational bottleneck objective (recovering that full objective would additionally require explicit optimization of $\mathcal R_1$). Nonetheless, if we treat $Z_2$ as the target label and $Z_1$ as the learned representation, then the JEPA objective mirrors the deep variational information bottleneck \citep{alemi2016deepVarIB}: it rewards representations that retain information useful for predicting the target through the $I(Z_1;Z_2)$ term, while penalizing information retained about the inputs through the compression terms. In our setting, this compression naturally appears symmetrically across the two input branches, through $\widehat I(X_1;Z_1)$ and $\widehat I(X_2;Z_2)$.

\subsection{Information Bottleneck in JEDI\'s Latent Conditional Diffusion}
The same perspective extends to latent conditional diffusion. This connection is especially natural from the PGM perspective underlying DDPM \citep{ho2020ddpm}, extended to the latent space (\cref{fig:joint-embedding-diffusion-pgm}). In JEDI, one-step latent prediction is replaced by a latent conditional denoising process that predicts the clean future latent ($z_2^0$) while conditioning on the current latent ($z_1^0$) and noised future latent ($z_2^t$). Following standard DDPM training, we optimize a conditional denoising objective consisting of an endpoint loss term $\mathcal L^0$ and denoising KLs:
\begin{equation}
\begin{aligned}
\mathcal L_{\mathrm{CDJ}}^{\mathrm{den}}
&:=
\mathcal L^0
+
\E_{p(x_1,x_2)}
\sum_{t=2}^{T}
\E_{q_\varphi(z_1^0\mid x_1)q_\varphi(z_2^0\mid x_2)q(z_2^t\mid z_2^0)}
\KL\left(
q(z_2^{t-1}\mid z_2^t,z_2^0)
\,\middle\|\,
p_\psi(z_2^{t-1}\mid z_2^t,z_1^0)
\right)
\\
\end{aligned}
\label{eq:main_cdj_denoising_loss}
\end{equation}
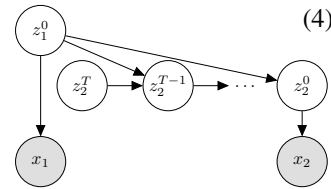
\begin{wrapfigure}{r}{0.3\textwidth}
    \vspace{-25pt}
    \centering
    \tikzset{latent/.style={circle, draw, minimum width=1cm, minimum height=1cm, inner sep=0pt, align=center}}
    \resizebox{\linewidth}{!}{%
    \begin{tikzpicture}[node distance=1.35cm]
    \node[obs]                      (x1) {$x_1$} ;
    \node[obs, right=4.2cm of x1]  (x2) {$x_2$} ;
    \node[latent, above=1.6cm of x1]   (z10) {$z_1^0$} ;
    \node[latent, above=0.5cm of x2]   (z20) {$z_2^0$} ;
    \node[left=0.3cm of z20]      (dots2) {$\ldots$} ;
    \node[latent, left=0.7cm of dots2]   (z2tm1) {$z_2^{T-1}$};
    \node[latent, left=0.7cm of z2tm1]   (z2t) {$z_2^{T}$};
    \edge {z10} {x1} ;
    \edge {z10} {z2tm1} ;
    \edge {z10} {z20} ;
    \edge {z20} {x2} ;
    \edge {z2t} {z2tm1} ;
    \edge {z2tm1} {dots2} ;
    \end{tikzpicture}
    }
    \caption{Latent conditional diffusion PGM. The clean latent $z_1^0$ acts as the context representation, and the reverse diffusion chain predicts the clean target latent $z_2^0$ through a denoising trajectory.}
    \label{fig:joint-embedding-diffusion-pgm}
\end{wrapfigure}
where the loss is typically implemented as regressing towards sampled noise \citep{ho2020ddpm}. This yields an analogous information-bottleneck structure (Appendix \ref{app:conditional-diffusion-jepa-ib}):
\begin{equation}
-\mathcal L_{\mathrm{CDJ}}^{\mathrm{den}}
\le
I(Z_1^0;Z_2^0)
-
\widehat I(X_1;Z_1^0)
-
\widehat I(X_2;Z_2^0)
+
\mathcal R_1
+
\mathcal R_{2,T}
-
\mathcal G
+
\mathcal C_2,
\label{eq:main_cdj_ib_den}
\end{equation}
where $\mathcal R_{2,T}$ is the terminal prior-matching term, and $\mathcal C_2$ is a constant with deterministic encoders. In our current implementation, we optimize this denoising term rather than the full objective augmented with $\mathcal R_1$. This perspective also points to a natural extension: explicitly regularizing the bottleneck term, potentially in ways related to recent collapse-prevention methods such as VICReg and LeJEPA \citep{shwartz2023informationVICREG, balestriero2025lejepa} (\cref{app:optimizing-the-bottlenck-regularizer,app:relation-vicereg-lejepa}).

This is the central mathematical motivation for JEDI: latent conditional denoising is not merely a generative surrogate, but supplies the predictive term in a JEPA-style variational bottleneck decomposition. Accordingly, our claim is that diffusion denoising provides a principled end-to-end predictive learning signal for the latent encoder, rather than that the method exactly optimizes the complete variational objective. 

\section{Method}
\label{sec:method_jedi}


\subsection{JEDI: Joint Embedding DIffusion World Model}
We introduce the Joint Embedding DIffusion World Model (JEDI), which learns an end-to-end latent space encoder while leveraging a diffusion dynamics model. The encoder compresses the observation into a clean latent $z_t^0$, the dynamics model predicts the next latent through conditional denoising, and reward and termination are predicted from the latent state. The world model is
\begin{equation}
\textbf{World Model:}
    \begin{cases}
        \text{Encoder:} & z_t^0 = \mathbf{E}_{\phi}(x_t), \\
        \text{Latent diffusion dynamics:} & \hat{z}_{t+1}^0 \sim \mathbf{S}(\mathbf{D}_{\theta}(\hat{z}_{t+1}^\tau,Z_t^\tau)),\\
        \text{Reward and termination:} & (\hat{r}_t, \hat{d}_t) = \mathbf{R}_{\psi}(z_t^0),
    \end{cases}
\label{eq:models}
\end{equation}
where $t$ is the environment time step, $x_t \in [0,1]^{64\times64\times3}$ is the observation, $z_t^0 \in [-3,3]^{16\times8\times8}$ is the clean latent, and $\tau$ is the diffusion time step sampled from a log-normal distribution that determines the noise scale. We write $Z_t^\tau := (c_{\mathrm{noise}}^\tau, z^0_{t-3:t}, a_{t-3:t})$ for the conditioning tuple consisting of the diffusion-time embedding, past latent states, and past actions. As in EDM \citep{karras2022EDM}, we use the preconditioned denoiser
\begin{equation}
\mathbf{D}_{\theta}(z_{t+1}^\tau, Z_t^\tau)=c_{\mathrm{skip}}^\tau z_{t+1}^\tau+c_{\mathrm{out}}^\tau\mathbf{F}_{\theta}\bigl(c_{\mathrm{in}}^\tau z_{t+1}^\tau, Z_t^\tau\bigr).
\end{equation}
Here $c_{\mathrm{noise}}^\tau$ is a fixed transformation of $\tau$, while $c_{\mathrm{skip}}^\tau$, $c_{\mathrm{out}}^\tau$, and $c_{\mathrm{in}}^\tau$ are the standard EDM preconditioning coefficients that improve the behavior of the neural network $\mathbf{F}_{\theta}$. Training and inference follow the same conditioning pipeline shown in \cref{fig:jedi_training_inference}. During training, the encoder first derives $z_{t-3:t+1}^0$ from observations $x_{t-3:t+1}$. The future latent $z_{t+1}^\tau$ is then obtained by perturbing $z_{t+1}^0$ with diffusion noise at level $\tau$, and $\mathbf{F}_{\theta}$ is trained to predict the direction toward the clean next latent. During inference, the conditioning tuple is formed in the same way, except that the future latent is initialized from noise and the solver $\mathbf{S}$ iteratively denoises it to obtain $\hat{z}_{t+1}^0$.

The key training signal is the joint embedding diffusion loss
{\setlength{\abovedisplayskip}{4pt}%
 \setlength{\belowdisplayskip}{6pt}%
 \setlength{\abovedisplayshortskip}{0pt}%
 \setlength{\belowdisplayshortskip}{0pt}%
\begin{equation}
\begin{split}
\E_{z_{1:T} \sim q, x_{1:T}\sim p, \tau \sim \textnormal{LN}}\Bigg[
\left\| \sum_{t=1}^{T}
\mathbf{F}_{\theta}\bigl(c_{\mathrm{in}}^\tau \mathrm{sg}(z_{t+1}^\tau), Z_t^\tau\bigr)
-
\frac{1}{c_{\mathrm{out}}^\tau}\bigl(\mathrm{sg}(z_{t+1}^0) - c_{\mathrm{skip}}^\tau\, \mathrm{sg}(z_{t+1}^\tau) \bigr)
\right\|^2
\Bigg].
\end{split}
\label{eqn:jedi_dyn_loss}
\end{equation}
}
where $q(z_t)$ denotes the deterministic pushforward of the observation distribution through the encoder, and $\mathrm{sg}$ denotes stop-gradient.
The reward and termination head is trained with
{\setlength{\abovedisplayskip}{4pt}%
 \setlength{\belowdisplayskip}{4pt}%
 \setlength{\abovedisplayshortskip}{0pt}%
 \setlength{\belowdisplayshortskip}{0pt}%
\begin{equation}
\begin{split}
\E_{z_{1:T}^0 \sim q, (x_{1:T},r_{1:T},d_{1:T}) \sim p}
\sum_{t=1}^T \textnormal{CE}(\mathbf{R_\psi}(z_t^0), (r_t,d_t)).
\label{eqn:jedi_rew_loss}
\end{split}
\end{equation}
}
Crucially, gradients from both losses update the encoder directly through $z_t^0$, so the latent is learned through predictive world-model training rather than through reconstruction.

\subsection{Practical Design Choices}
Several design choices are important in practice. First, to reduce representation collapse, we apply stop-gradient to the future latent target and set the encoder learning rate to 0.3 times the denoiser learning rate, following the spirit of JEPA and TDMPC2 \citep{assran2023I-JEPA,hansen2023tdmpc2}. This allows the encoder to learn directly from the denoising objective while preventing unstable feedback through the future latent target. Second, to stabilize iterative denoising in latent space, we clamp latent activations with the differentiable transform $C(z)=\tanh(z/s)s$ with $s=3$, applied to both encoder outputs and denoiser outputs. This is common practice: DIAMOND clamps pixels to the natural $[-1,1]$ image range and discretizes that range into 256 bins, while HI uses a plain $\tanh$ to clamp latents \citep{alonso2024diamond,cohen2026horizonImagination}. Third, with every trajectory batch we randomly switch with uniform probability between using the denoiser output $\mathbf{D}_{\theta}$ and the encoder output $\mathbf{E}_{\phi}$ as the next-step conditioning signal. This balances near-horizon consistency with sufficient direct encoder supervision.

The policy is trained in an actor-critic framework \citep{hafner2020Dreamerv2,alonso2024diamond}. We intentionally use REINFORCE \citep{williams1992REINFORCE} so that the actor does not require backpropagation through multi-step denoising, which keeps policy learning lightweight despite the iterative diffusion world model.

\section{Experiments}
\label{sec:experiments}

\begin{figure}[t]
    \centering
    \begin{subfigure}[t]{0.67\linewidth}
        \centering
        \includegraphics[width=\textwidth]{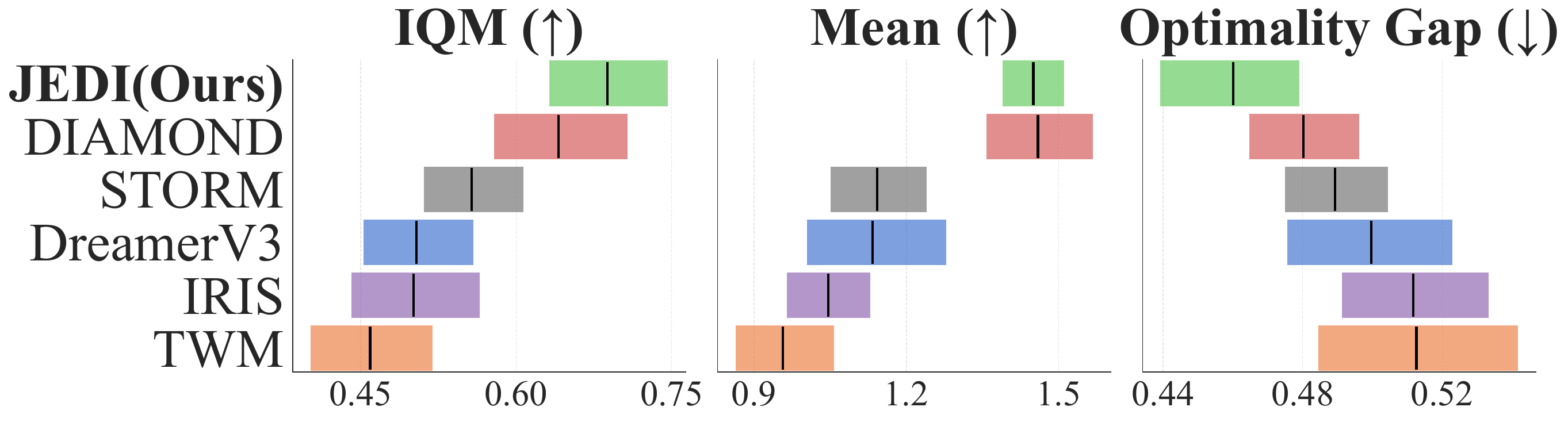}
    \end{subfigure}
    \hfill
    \begin{subfigure}[t]{0.28\linewidth}
        \centering
        \includegraphics[width=\textwidth]{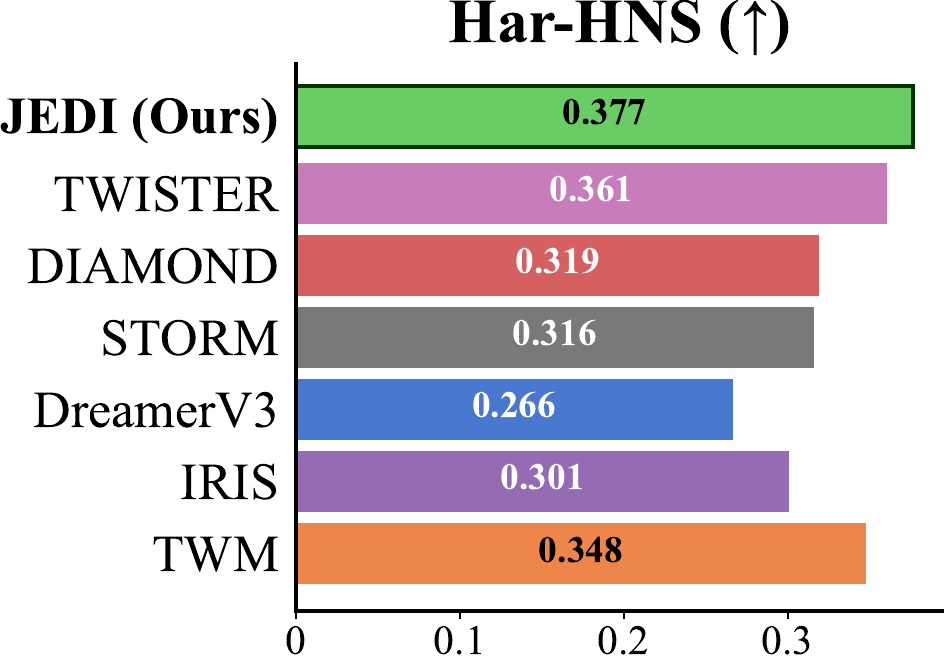}
    \end{subfigure}
    \caption{Aggregate Atari100k performance. Left: IQM, mean, and optimality gap following \citet{agarwal2021precipice}. Right: Har-HNS (\cref{sec:harhns_discussion}) is shown as a complementary descriptive statistic because JEDI exhibits a different task-level profile from prior baselines. TWISTER does not report seed-level results, so mean, IQM and optimality-gap error bars are unavailable. Full per-game numbers are reported in Appendix \cref{tab:atari100k_FULL_NUMBERS}.}
    \label{fig:main_result}
\end{figure}
\begin{figure*}[t]
    \centering
    \includegraphics[width=0.8\textwidth]{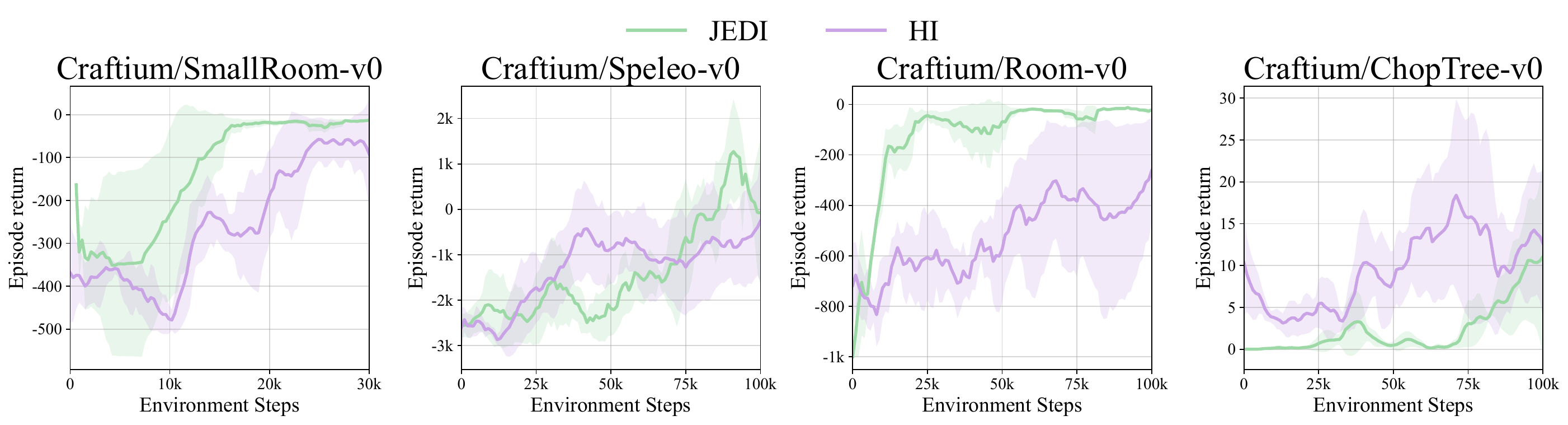}
    \caption{Craftium performance comparison across JEDI and HI}
    \label{fig:craftium_JEDI_HI}
\end{figure*}

\paragraph{Experimental Setup}
Our primary baselines are DIAMOND \citep{alonso2024diamond} and HI \citep{cohen2026horizonImagination}. DIAMOND is the canonical online pixel-space diffusion world model and the cleanest comparison for the central pixel-versus-end-to-end-latent question. HI is the most relevant latent-diffusion comparison, since it shows that latent diffusion can be efficient but does not learn the latent space end to end through the predictive world-model objective. We also compare against DreamerV3, STORM\footnote{We derive the latest updated STORM results from \cite{OC-STORM}}, TWM, IRIS, and TWISTER when published numbers are available \citep{hafner2023Dreamerv3, zhang2023storm, OC-STORM, robine2023TWM, micheli2022iris, TWISTER}. 

We evaluate JEDI primarily on the full Atari100k benchmark with 26 tasks and 5 seeds per task \citep{kaiser2019Atari100kSimPLE}. Each run takes about two days on an A100. For each task, we report the better result from two diffusion-model batch sizes, 32 and 64. Following \citet{agarwal2021precipice}, we emphasize interquartile mean, optimality gap, performance profile of the Human-Normalized Scores (HNS), and we use bootstrap sampling for error bars when seed-level data are available. Refer to \cref{appendix:benchmark-settings} for more details.
\begin{wrapfigure}{r}{0.4\textwidth}
    \centering
    \includegraphics[width=\linewidth]{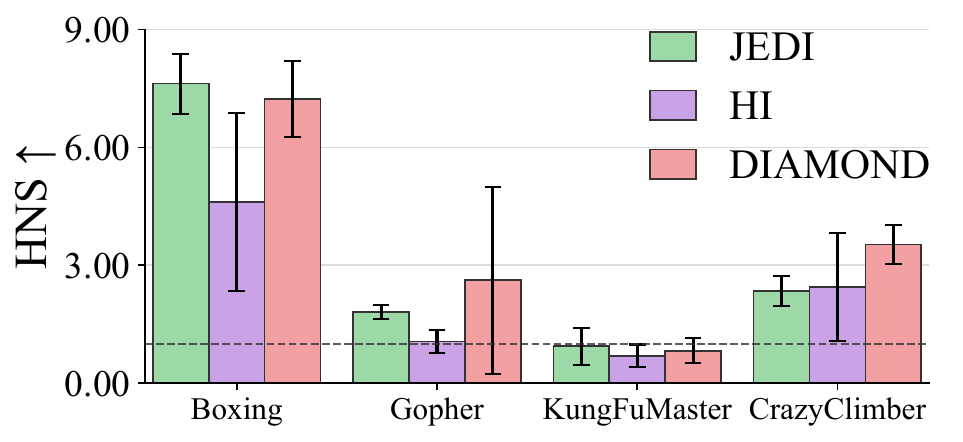}
    \caption{Atari100k performance comparison across JEDI, HI, and DIAMOND}
    \label{fig:Atari_JEDI_HI_DIAMOND}
    \vspace{-20pt}
\end{wrapfigure}

In addition to Atari100k, we evaluate on Craftium \citep{malagon2024craftium}, a 3D embodied Minecraft-like decision-making environment that lets us test whether JEDI transfers beyond 2D Atari. For Craftium, we run 3 seeds per task. Craftium also gives the clearest direct empirical comparison to HI, since it reports published training-curve results there but not on the full Atari100k benchmark.

\paragraph{Har-HNS}
\label{sec:harhns_discussion}
Arithmetic mean HNS is an important aggregate metric, but it is disproportionately influenced by already high-valued games and can therefore obscure large relative gains on low-HNS tasks \citep{how_not_to_lie_with_stats, joint2008CompositeIndicators}. As a result, methods that improve low-HNS games can appear less impressive under arithmetic averaging than their task-level profile would suggest. Critically, although JEDI and DIAMOND have similar mean HNS, they improve in very different HNS regimes (\cref{fig:task_performance_profile,fig:hns_performance_profile}). 

Inspired by the F1 score \citep{van1979information}, the use of harmonic mean provides a complementary perspective by emphasizing improvements where performance is still poor. Formally, we define
\begin{equation}
\mathrm{Har\text{-}HNS}
=
\left(
\frac{1}{N}
\sum_{i=1}^{N}
\frac{1}{\mathrm{HNS}_i + 0.1}
\right)^{-1},
\end{equation}
where $N$ is the number of games, $\mathrm{HNS}_i$ is the average HNS for game $i$, and the offset of $0.1$ ensures numerical stability for negative HNS values. Intuitively, mean HNS asks how much total score we collect across tasks, whereas Har-HNS asks how well we are doing on the tasks that remain very difficult. We therefore use Har-HNS as a complement to mean HNS to better understand the changed performance profile induced by end-to-end latent diffusion.

\subsection{Main Atari100k Results}
\label{sec:results}
As shown in \cref{fig:main_result} and the full per-game breakdown in Appendix \cref{tab:atari100k_FULL_NUMBERS}, JEDI achieves strong Atari100k performance. It matches or exceeds strong baselines on IQM, optimality gap, and number of state-of-the-art game scores, while remaining competitive on arithmetic mean HNS and achieving runner-up performance in the number of superhuman tasks. 
\begin{wrapfigure}[17]{r}{0.45\textwidth}
    \centering
    \includegraphics[width=\linewidth,
                   trim=3pt 2pt 3pt 2pt,clip]{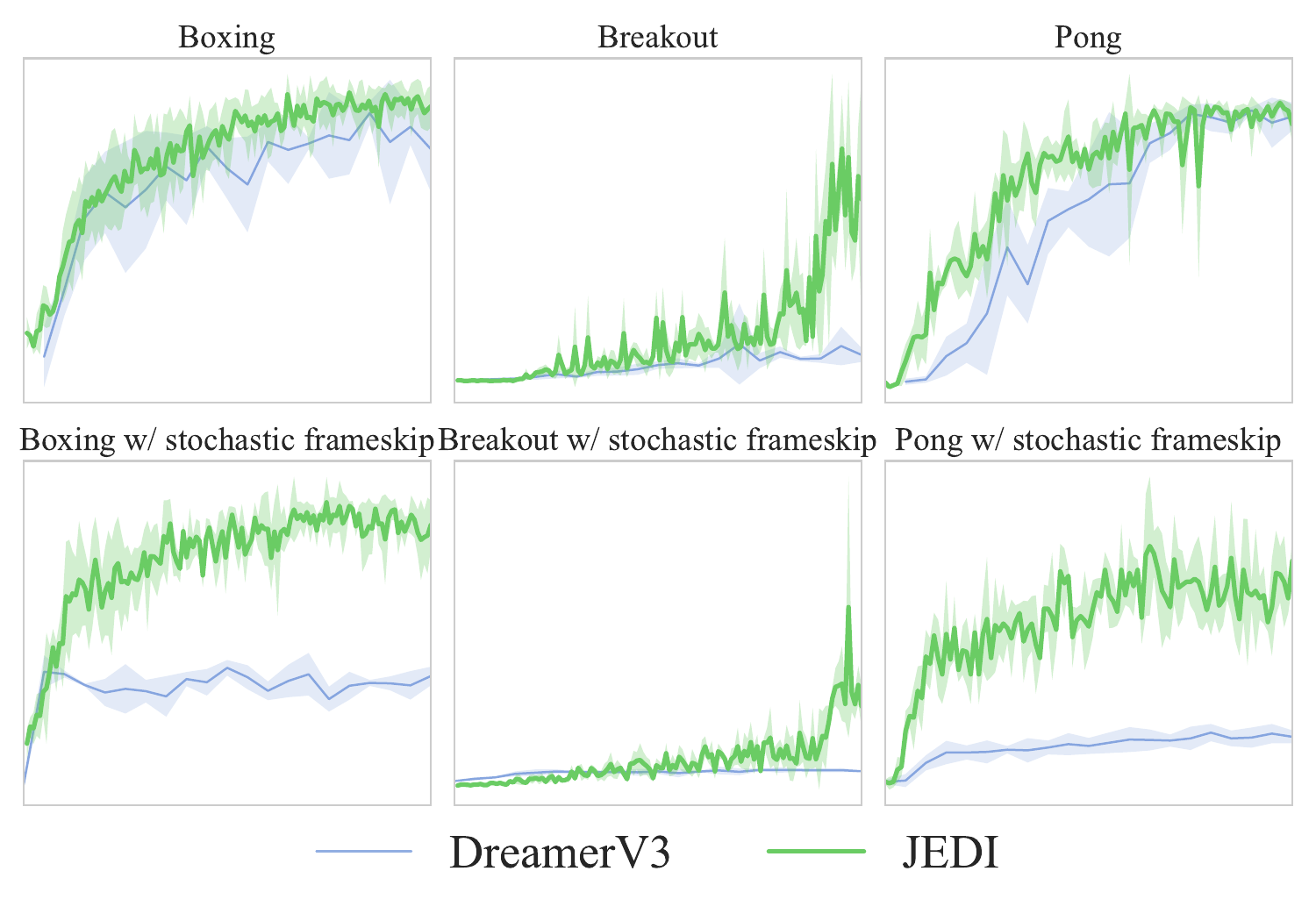}
    \caption{\small JEDI vs. DreamerV3 on Atari with random frameskip. The top row shows performance without frame-skipping, and the bottom row shows performance with stochastic frame-skipping.}
    \vspace{-80pt}
    \label{fig:stoch_experiments}
\end{wrapfigure}

On the four Atari tasks reported by HI, JEDI also performs better overall on the shared subset (\cref{fig:Atari_JEDI_HI_DIAMOND}). This sharpens the paper's main empirical point. The contribution is not merely that latent diffusion can be cheaper than pixel diffusion, but that an end-to-end predictive denoising, trained directly through the world-model objective, is sufficient to support a competitive online diffusion world model without the need for a pretrained model for learning latents.

\paragraph{Stochastic Atari.} All of Atari100k tasks are deterministic. To test whether diffusion helps model stochastic targets, we compare JEDI with DreamerV3, an MLP-based end-to-end latent world model, on three Atari environments with random frameskips between 2 and 6. JEDI consistently outperforms DreamerV3, suggesting that diffusion may be more robust to high task-related aleatoric uncertainty.

\subsection{Craftium Results}
Craftium provides supporting evidence that JEDI is not specific to Atari100k. As shown in \cref{fig:craftium_JEDI_HI}, the same qualitative picture appears here as well: end-to-end predictive latents can support a strong latent diffusion world model without reconstruction-based latent training. In a direct comparison with HI on four Craftium tasks, JEDI performs better overall. Because Craftium is a 3D embodied Minecraft-like environment, these results also suggest that the approach extends beyond 2D Atari.

\subsection{Efficiency and Scalability}
\begin{figure}[t]
  \centering
  \begin{subfigure}[t]{0.3\linewidth}
    \centering
    \includegraphics[width=\textwidth]{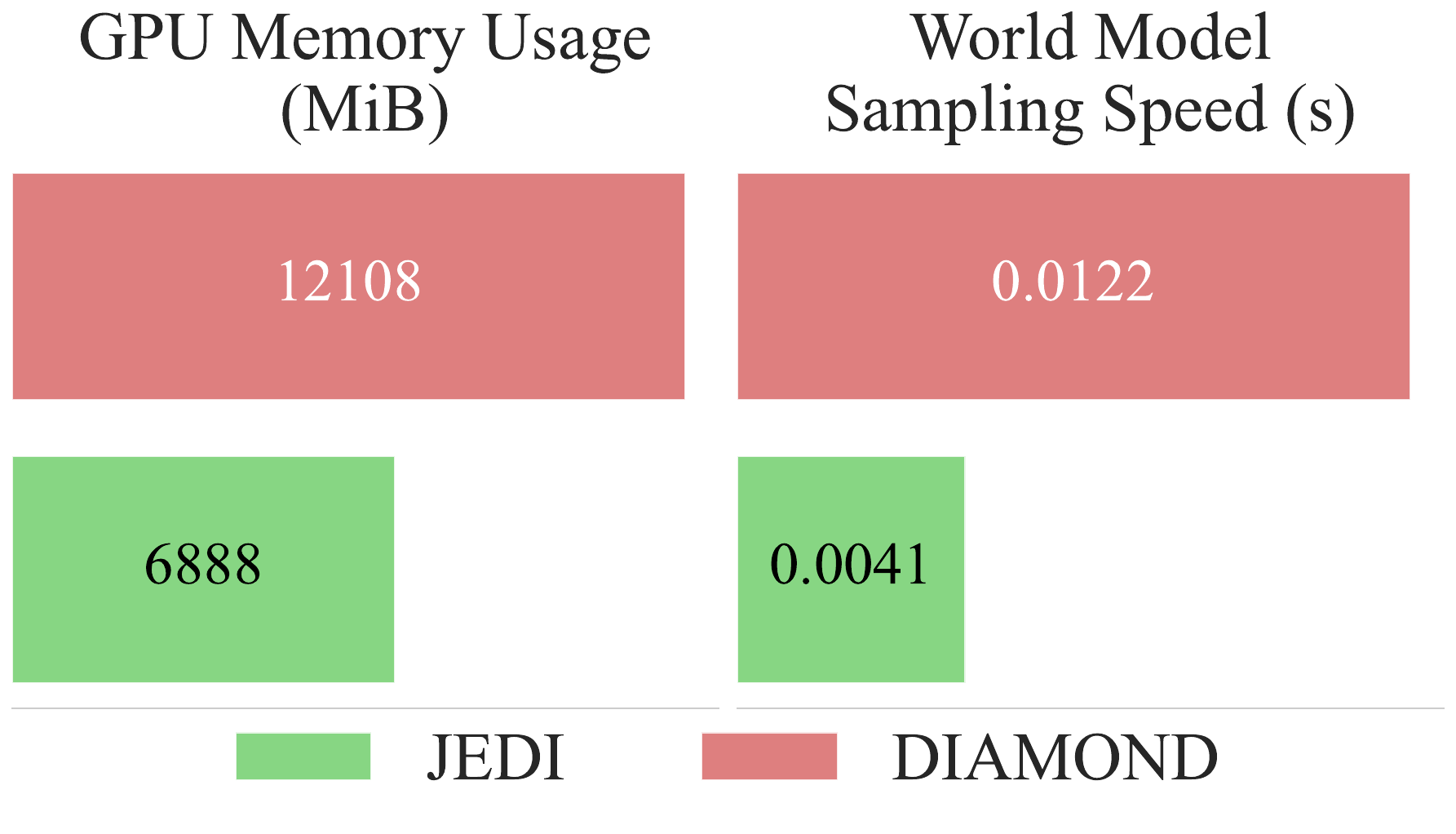}
  \end{subfigure}
    \hspace{0.1\textwidth}
  \begin{subfigure}[t]{0.45\linewidth}
    \centering
    \includegraphics[width=\textwidth]{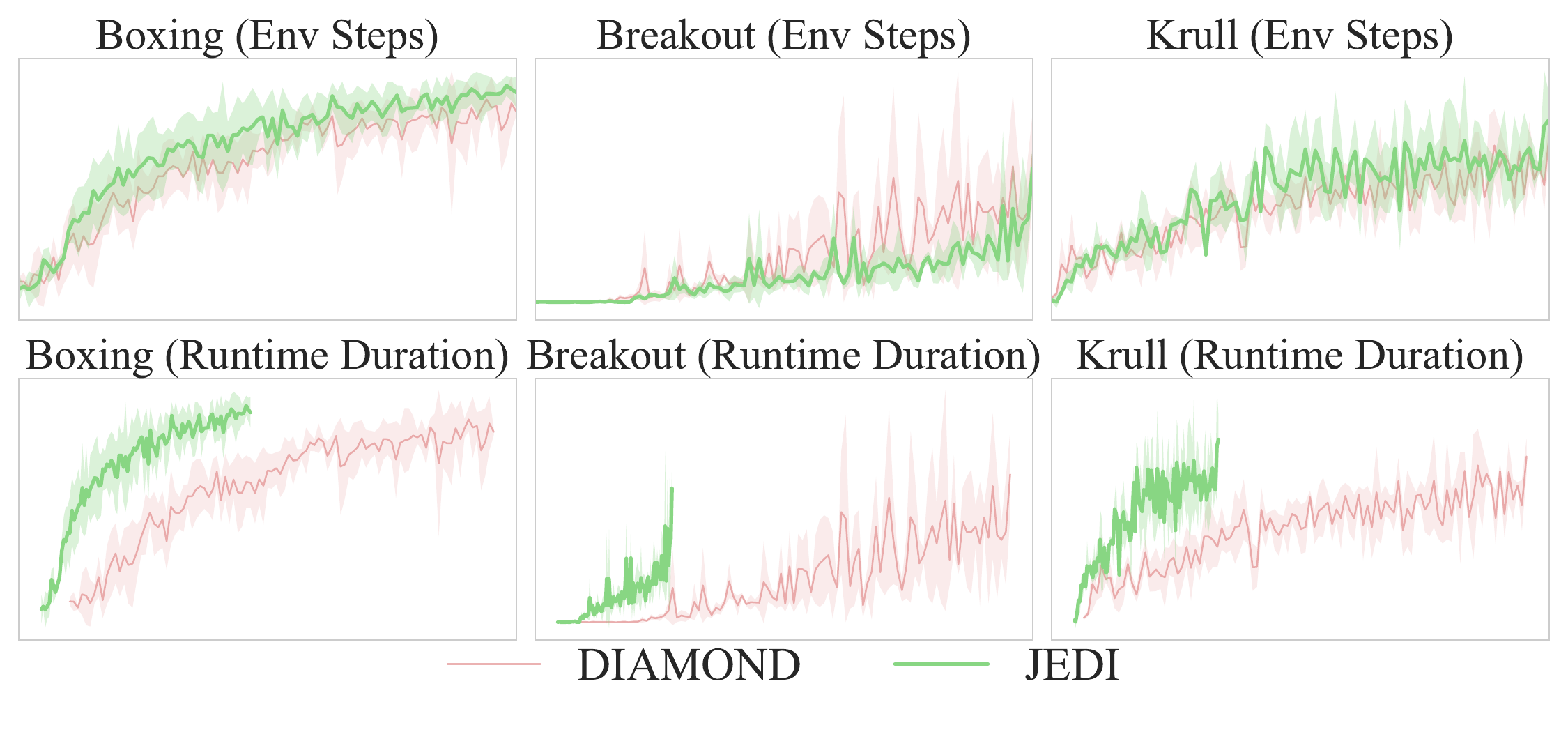}
  \end{subfigure}
  \caption{Left: JEDI uses 57\% of DIAMOND's GPU memory while sampling the world model over 3$\times$ faster. Right: JEDI reaches comparable performance in less than half the wall-clock training time.}
  \label{fig:compute_and_runtime_comparison}
\vspace{-10pt}
\end{figure}
\begin{wraptable}{r}{0.45\textwidth}
\vspace{-35pt}
\centering
\scriptsize
\begin{tabular}{lccc}
\hline
Method & A100 (hrs) & V100 (hrs) & \#Params\\
\hline
JEDI      & \textcolor{pastelgreen}{38} & -   & \textcolor{pastelgreen}{13.5M} \\ 
HI \citep{cohen2026horizonImagination}        & \textcolor{pastelpurple}{27}   & -   & \textcolor{pastelpurple}{97M} \\
DIAMOND \citep{alonso2024diamond}   & \textcolor{pastelred}{98}   & -   & \textcolor{pastelred}{13.5M} \\
DreamerV3 \citep{hafner2023Dreamerv3} & -                           & 12  & 18M\\
STORM \citep{zhang2023storm}    & 7                           & 9.3 & 18.8M \\
TWM   \citep{robine2023TWM}    & 10 & - & 21.6M\\
IRIS \citep{micheli2022iris}      & 168 & - & 30M\\

\hline
\end{tabular}
\caption{Wall clock run times per Atari100k run on A100 / V100 and model size}
\label{tab:wall_clock_runtime}
\vspace{-15pt}
\end{wraptable}
As summarized in \cref{fig:compute_and_runtime_comparison}, relative to DIAMOND, JEDI uses 43\% less VRAM, achieves more than 3$\times$ faster world-model sampling, and trains more than 2$\times$ faster. These gains come from moving diffusion from an observation in $\mathbb{R}^{64\times 64 \times 3}$ to a 12$\times$ smaller $z_t^0 \in \mathbb{R}^{16\times 8 \times 8}$ compact learned latent with end-to-end training. This is the clearest practical takeaway of the paper: predictive latent diffusion is much more scalable than pixel-space diffusion for online MBRL.

The most relevant efficiency comparison is DIAMOND: as shown in \cref{tab:wall_clock_runtime}, JEDI stays competitive while cutting A100 training time from 98 to 38 hours at the same parameter count. HI is faster, but JEDI better preserves latent diffusion's efficiency advantage without giving up competitive end-to-end performance.

\subsection{Ablation Study}
We conduct two sets of ablation studies, each over five Atari tasks. The first set studies how the latent representation is learned, while the second isolates several stabilization and design choices.
\begin{wrapfigure}{r}{0.4\textwidth}
    \vspace{-10pt}
    \centering
    \includegraphics[width=\linewidth]{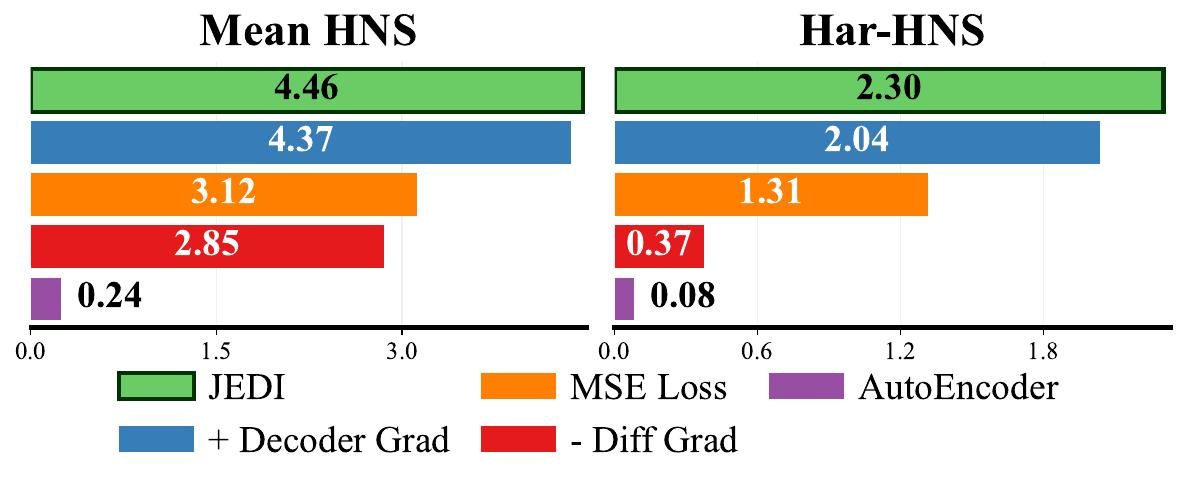}
    \caption{Latent-learning ablations on five Atari tasks. Joint predictive training outperforms adding decoder-based reconstruction supervision (+ Decoder Grad), substituting diffusion loss for MSE loss (MSE Loss), removing diffusion loss gradients (- Diff Grad) and using separately VAE trained latents (AutoEncoder).}
    \label{fig:latent_ablation}
    \vspace{-40pt}
\end{wrapfigure}
\paragraph{Latent Learning ablation}
\Cref{fig:latent_ablation} shows that how the latent space is learned matters. Full JEDI performs best. Replacing the denoising objective with direct next-state MSE prediction reduces performance, and removing diffusion-loss gradients hurts further. Using a separately trained VAE latent performs worst, while adding decoder gradients also slightly hurts performance.

Overall, these ablations support learning the latent space jointly with the predictive world model and point to the denoising objective as a key training signal.
\begin{wrapfigure}{r}{0.45\textwidth}
    \vspace{-75pt}
    \centering
    \includegraphics[width=\linewidth]{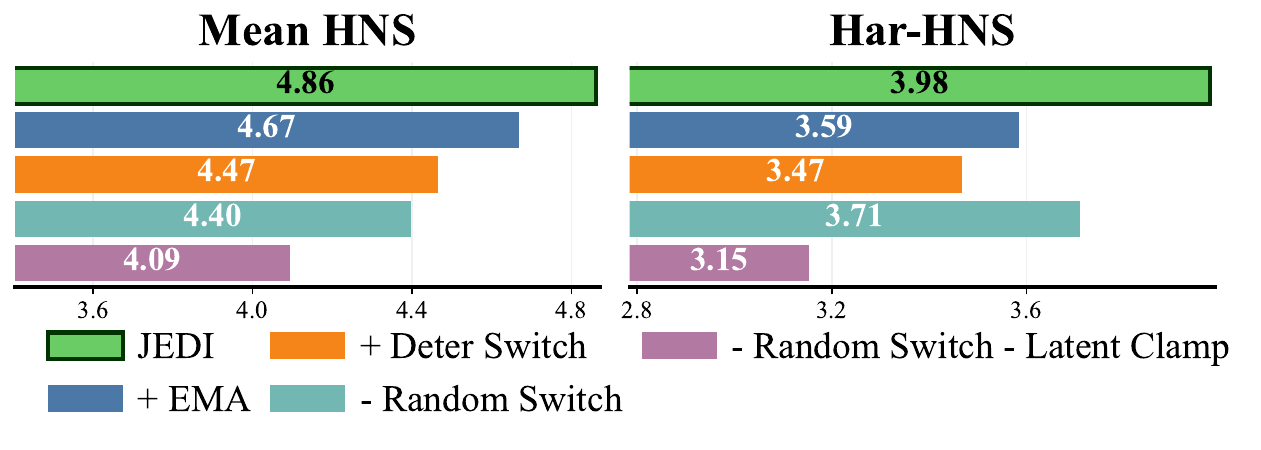}
    \caption{Design-choice ablations. Variants test EMA targets, deterministic switching, removing random switching, and removing switching together with latent clamping.}
    \label{fig:design_ablation}
    \vspace{-5pt}
\end{wrapfigure}
\paragraph{Design Choice Ablation}
We also study the design choices that help make end-to-end latent diffusion stable and competitive (\cref{fig:design_ablation}). Variants using an EMA target encoder, deterministic switching, or no random switching perform worse than the default design, and removing switching together with latent clamping hurts further. Still, HNS remains high across these ablations, suggesting that these design choices are helpful rather than critical. The end-to-end predictive objective appears to be the main ingredient, while the stabilization techniques remain important supporting details.

\section{Discussion}
\label{sec:discussion}
\subsection{Performance Profile and Har-HNS}
\begin{figure*}[t]
    \centering
    \includegraphics[width=0.85\textwidth]{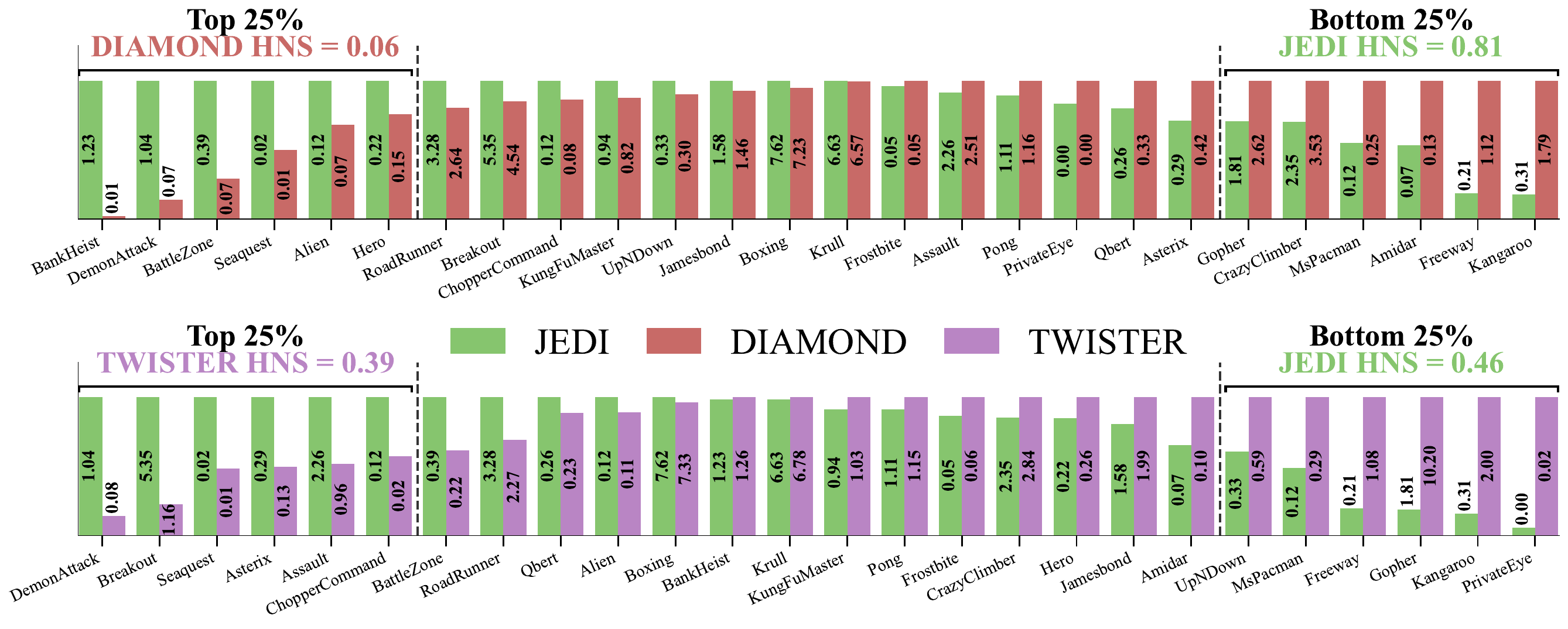}
    \caption{Task-level performance profiles for JEDI versus DIAMOND and TWISTER. Tasks are ordered by JEDI's relative gain over each comparison method, with the top and bottom quartiles shown separately. Bar heights are normalized, while the annotated values report HNS. JEDI excels against DIAMOND on low-HNS tasks, whereas DIAMOND tends to outperform JEDI on high-HNS tasks. This distinct performance profile is less apparent when comparing JEDI with TWISTER, a latent baseline.}
    \label{fig:task_performance_profile}
\end{figure*}

JEDI has a distinctly different performance profile from DIAMOND (\cref{fig:task_performance_profile}). Importantly, we build directly on top of DIAMOND and keep the experimental setup as close as possible, with the main change being the end-to-end JEPA latent space and only minor hyperparameter adjustments (\cref{appendix:hyperparameters}). JEDI's top quantile of relative gains occurs where DIAMOND's HNS is very low (mean $\approx 0.06$), whereas DIAMOND's top quantile occurs in much higher-HNS regimes (mean $\approx 0.81$), as shown in \cref{fig:task_performance_profile,fig:hns_performance_profile}. 

This accounts for \cref{fig:main_result} the aggregate ranking flips under Har-HNS (\cref{sec:harhns_discussion}): although JEDI ($1.450$) is slightly below DIAMOND ($1.459$) on arithmetic mean HNS (\cref{tab:atari100k_FULL_NUMBERS}), it attains the strongest Har-HNS at $0.377$, ahead of DIAMOND ($0.319$). This indicates that JEDI fails less severely on the games that remain hardest. 
\subsection{Why Does the Performance Profile Change?}

\begin{wrapfigure}{r}{0.4\textwidth}
    \vspace{-40pt}
    \centering
    \includegraphics[width=\linewidth]{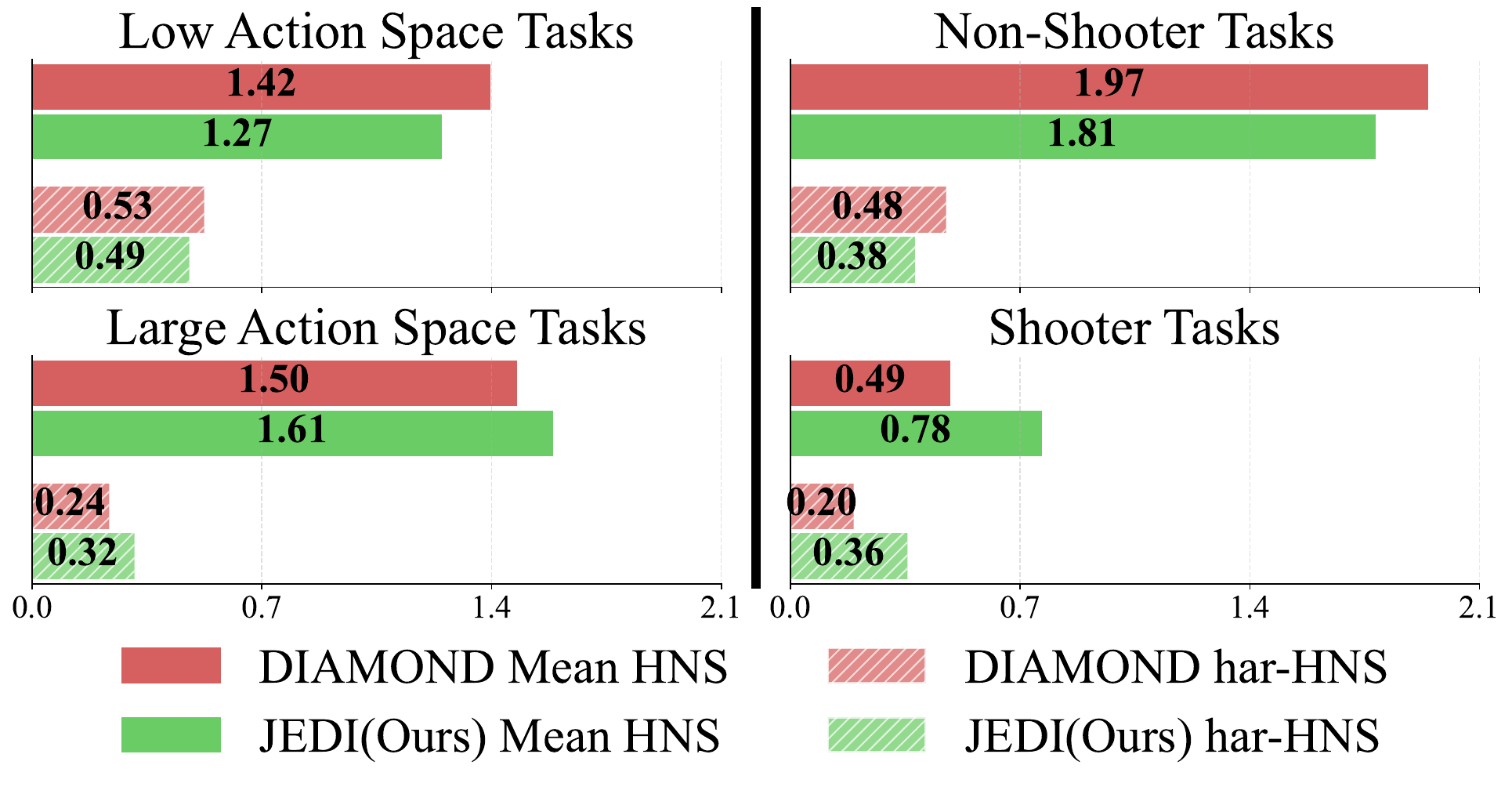}
    \caption{Task-property analysis comparing high versus low action-space games and shooter versus non-shooter games.}
    \label{fig:action_shooter_comparison}
    \vspace{-10pt}
\end{wrapfigure}
A plausible explanation is that introducing an end-to-end latent interface changes which games are easiest for the policy to optimize. All of JEDI's six top-quantile games have shooter-style dynamics with five of them having the maximum action space, motivating the aggregate breakdown in \cref{fig:action_shooter_comparison} (see also \cref{tab:tasks_with_num_actions_shooter_task}). As summarized in \cref{fig:action_shooter_comparison}, JEDI's largest gains are concentrated in higher-complexity games, especially games with larger action spaces and shooter-style dynamics. DIAMOND, in contrast, feeds raw image features to its actor and critic, so under a fixed compute budget it must spend capacity on both representation learning and policy/value learning. JEDI instead gives the actor and critic a learned latent $z_t^0 \in \mathbb{R}^{16\times 8 \times 8}$ rather than an observation in $\mathbb{R}^{64\times 64 \times 3}$, making the actor-critic input about 12$\times$ smaller and reducing the need for those networks to solve representation learning themselves. 

Intuitively, the effective interaction space between inputs and action/value prediction can grow combinatorially---often effectively exponentially---with input dimension, so this reduction may matter most in large-action, visually complex games. We regard this as an interpretation supported by the empirical pattern, not as a formal causal proof.

\subsection{Qualitative Analysis}
\label{sec:qualitative_analysis}
\begin{figure*}[t]
    \centering
    \includegraphics[width=0.8\textwidth]{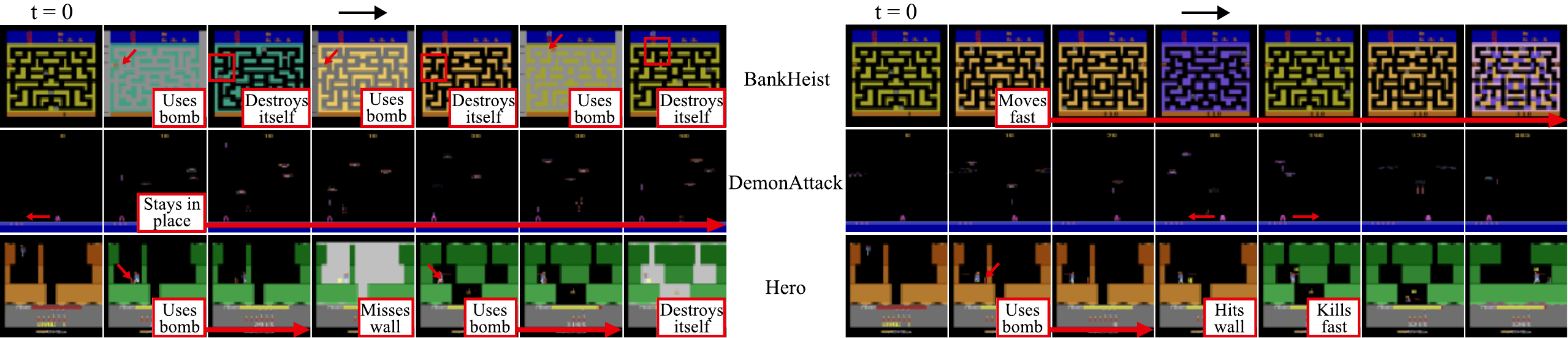}
    \caption{Example trajectories with DIAMOND (\textbf{LEFT}) and JEDI (\textbf{RIGHT}) on three tasks. JEDI is significantly more effective at eliminating enemies and obstacles while effectively minimizes self-destruction as compared to DIAMOND}
    \label{fig:qualitative-analysis}
\end{figure*}

The changed task-level performance profile is also visible qualitatively (\cref{fig:qualitative-analysis}). On \textit{BankHeist}, \textit{DemonAttack}, and \textit{Hero}---all in JEDI's top relative-performance quantile---JEDI learns visibly different policies from DIAMOND: it repeatedly re-enters maps in \textit{BankHeist} to collect easy bank spawns while avoiding self-destructive \textit{FIRE} actions, more reliably destroys obstacles in \textit{Hero}, and actively tracks and neutralizes enemies in \textit{DemonAttack} rather than staying in a corner and firing ineffectively.
\setlength{\tabcolsep}{1pt}

We view these trajectories as consistent with the broader performance-profile story: moving to an end-to-end predictive latent may help more on complex games with large action spaces and shooter dynamics, where a compressed world-model latent may better align the policy-learning interface with downstream value estimation.

\paragraph{Conclusion}
To our knowledge, JEDI is the first to show that diffusion world models can be trained in an end-to-end predictive latent space using JEPA-style learning. JEDI preseres strong online MBRL performance and substantially improving efficiency over pixel diffusion. Beyond the immediate compute gains, our results suggest that predictive latent learning changes the behavior of the world model itself: JEDI exhibits a markedly different task-level performance profile from pixel-space diffusion, indicating that end-to-end latent learning affects more than runtime and memory alone. Together with the information-bottleneck perspective developed in our theory section, these findings position predictive latent diffusion as a principled and practical direction for scaling online MBRL.

\paragraph{Limitations.}
\label{sec:limitations}
Our evidence is limited to Atari100k (5 seeds) and four Craftium tasks (3 seeds; \cref{fig:main_result,fig:craftium_JEDI_HI}), so we have not yet tested JEDI on DMControl, Procgen, or higher-resolution domains. JEDI is also not the fastest latent method: HI trains in 27 A100-hours versus our 38, though HI uses a pretrained perceptual model and 97M parameters, versus JEDI's 13.5M (\cref{tab:wall_clock_runtime}). Theoretically, \cref{eq:jepa_loss,eq:main_jepa_ib,eq:main_jepa_ib_bound,eq:main_cdj_denoising_loss,eq:main_cdj_ib_den} motivate representation-learning properties rather than guaranteeing optimization. Performance also depends on several design choices, including tanh clamping and random switching (\cref{fig:latent_ablation,fig:design_ablation}); DIAMOND and HI also use clamping, and DIAMOND additionally uses latent discretization. Finally, performance is sensitive to denoiser batch size, so we report the better of batch sizes 32 and 64 for each task; HI similarly reports results across multiple hyperparameter settings.

\section{Related Works}
\label{sec:related}
\paragraph{World models and online MBRL.}
Classic approaches such as Dyna and PlaNet established predictive world models for control \citep{sutton1991DYNA,hafner2019Planet}. Search-based successors such as MuZero and EfficientZero are outside our scope because we focus on online MBRL without look-ahead planning \citep{silver2017MuZero,muzero2020,EfficientZero}. Dreamer and its successors showed that actor-critic learning on imagined trajectories scales to visual domains \citep{Hafner2020Dreamer,hafner2020Dreamerv2,hafner2023Dreamerv3}. IRIS, TWM, STORM, and TWISTER expanded the Atari100k design space with sequence-model world models \citep{micheli2022iris,robine2023TWM,zhang2023storm,TWISTER}. DIAMOND introduced strong pixel diffusion for online MBRL, while Horizon Imagination moved diffusion to a separately learned latent space \citep{alonso2024diamond,cohen2026horizonImagination}. Other concurrent methods add annotations, event supervision, or explicit exploration \citep{OC-STORM,EA-SIMULUS,SIMULUS} and are beyond our scope of studying end-to-end latent diffusion for MBRL world model backbone.
\paragraph{JEPA and predictive representations.}
Early work argued that representations should preserve predictive structure rather than reconstruct raw detail \citep{PMAX}, a view further developed by modern JEPA-style methods \citep{grill2020BYOL,caron2021DINO,assran2023I-JEPA,bardes2024vjepa,assran2025VJEPA2}. Related predictive-latent methods also perform well in model-based control \citep{hansen2022TDMPC,hansen2023tdmpc2,maes2026leworldmodel,wang2026temporalstraightening}. Recent work further connects JEPA to the information bottleneck \citep{tishby2000InfoBottleneck,alemi2016deepVarIB,shwartz2017opening,2023multiviewJepa,shwartz2023informationVICREG,shwartz2024compress}. To our knowledge, JEDI is the first online world model to combine a JEPA-style objective with diffusion denoising and an information-bottleneck interpretation.
\paragraph{Diffusion}
Diffusion models are now a standard generative framework \citep{song2019NCSN,ho2020ddpm,karras2022EDM}. Many high-capacity systems use separately learned latent spaces for efficiency \citep{rombach2022StableDiffusion,esser2024StableDiffusion3,bruce2024genie,parkerholder2024genie2,liu2024sora}. LSGM also studied end-to-end latent generative modeling, but with reconstruction-based training \citep{vahdat2021LSGM}. Diffusion has also been effective in planning, control, and offline world modeling \citep{janner2022diffuser,chi2023DiffusionPolicy,jackson2024pgd,wang2022DiffPolicyExpressive,frans2025diffusionGuidance,lu2023synther,ding2024diffusion}, and has been used for representation learning through denoising-step search, distillation, compressed conditioning, and hidden-feature extraction \citep{luo2023DiffHyperfeatures,yue2024ExpDiffTimeSteps,yang2023NUSDiffasRepLearner,zhang2022UnsupRepLearnPretrainedSEA,pandey2022diffusevae,nielsen2023diffenc,baranchuk2021LabelEfficientUNETFeat,xiang2023DiffAEAreUnified}. JEDI instead learns and predicts compressed latents end to end through a JEPA-style diffusion world-model objective.

\bibliography{references}

@article{balestriero2025lejepa,
  title={Lejepa: Provable and scalable self-supervised learning without the heuristics},
  author={Balestriero, Randall and LeCun, Yann},
  journal={arXiv preprint arXiv:2511.08544},
  year={2025}
}

@article{malagon2024craftium,
  title={Craftium: Bridging Flexibility and Efficiency for Rich 3D Single-and Multi-Agent Environments},
  author={Malag{\'o}n, Mikel and Ceberio, Josu and Lozano, Jose A},
  journal={arXiv preprint arXiv:2407.03969},
  year={2024}
}

@article{PMAX,
  title={Discovering predictable classifications},
  author={Schmidhuber, J{\"u}rgen and Prelinger, Daniel},
  journal={Neural Computation},
  volume={5},
  number={4},
  pages={625--635},
  year={1993},
  publisher={MIT Press}
}

@article{assran2025VJEPA2,
  title={V-jepa 2: Self-supervised video models enable understanding, prediction and planning},
  author={Assran, Mido and Bardes, Adrien and Fan, David and Garrido, Quentin and Howes, Russell and Muckley, Matthew and Rizvi, Ammar and Roberts, Claire and Sinha, Koustuv and Zholus, Artem and others},
  journal={arXiv preprint arXiv:2506.09985},
  year={2025}
}

@article{wang2026temporalstraightening,
  title={Temporal straightening for latent planning},
  author={Wang, Ying and Bounou, Oumayma and Zhou, Gaoyue and Balestriero, Randall and Rudner, Tim GJ and LeCun, Yann and Ren, Mengye},
  journal={arXiv preprint arXiv:2603.12231},
  year={2026}
}

@article{maes2026leworldmodel,
  title={Leworldmodel: Stable end-to-end joint-embedding predictive architecture from pixels},
  author={Maes, Lucas and Lidec, Quentin Le and Scieur, Damien and LeCun, Yann and Balestriero, Randall},
  journal={arXiv preprint arXiv:2603.19312},
  year={2026}
}

@article{lou2023discreteDiffLMSEDD,
  title={Discrete diffusion modeling by estimating the ratios of the data distribution},
  author={Lou, Aaron and Meng, Chenlin and Ermon, Stefano},
  journal={arXiv preprint arXiv:2310.16834},
  year={2023}
}

@article{prabhudesai2507diffusionBeatLM,
  title={Diffusion beats autoregressive in data-constrained settings},
  author={Prabhudesai, Mihir and Wu, Mengning and Zadeh, Amir and Fragkiadaki, Katerina and Pathak, Deepak},
  journal={arXiv preprint arXiv:2507.15857},
  year={2025}
}

@article{li2022diffusion-lm,
  title={Diffusion-lm improves controllable text generation},
  author={Li, Xiang and Thickstun, John and Gulrajani, Ishaan and Liang, Percy S and Hashimoto, Tatsunori B},
  journal={Advances in neural information processing systems},
  volume={35},
  pages={4328--4343},
  year={2022}
}

@article{frans2025diffusionGuidance,
  title={Diffusion guidance is a controllable policy improvement operator},
  author={Frans, Kevin and Park, Seohong and Abbeel, Pieter and Levine, Sergey},
  journal={arXiv preprint arXiv:2505.23458},
  year={2025}
}

@article{belhasin2025advancing,
  title={Advancing Image Classification with Discrete Diffusion Classification Modeling},
  author={Belhasin, Omer and Golan, Shelly and El-Yaniv, Ran and Elad, Michael},
  journal={arXiv preprint arXiv:2511.20263},
  year={2025}
}

@article{shwartz2023informationVICREG,
  title={An information theory perspective on variance-invariance-covariance regularization},
  author={Shwartz-Ziv, Ravid and Balestriero, Randall and Kawaguchi, Kenji and Rudner, Tim GJ and LeCun, Yann},
  journal={Advances in neural information processing systems},
  volume={36},
  pages={33965--33998},
  year={2023}
}

@article{alemi2016deepVarIB,
  title={Deep variational information bottleneck},
  author={Alemi, Alexander A and Fischer, Ian and Dillon, Joshua V and Murphy, Kevin},
  journal={arXiv preprint arXiv:1612.00410},
  year={2016}
}

@article{shwartz2017opening,
  title={Opening the black box of deep neural networks via information},
  author={Shwartz-Ziv, Ravid and Tishby, Naftali},
  journal={arXiv preprint arXiv:1703.00810},
  year={2017}
}

@article{tishby2000information,
  title={The information bottleneck method},
  author={Tishby, Naftali and Pereira, Fernando C and Bialek, William},
  journal={arXiv preprint physics/0004057},
  year={2000}
}

@article{shwartz2024compress,
  title={To compress or not to compress—self-supervised learning and information theory: A review},
  author={Shwartz Ziv, Ravid and LeCun, Yann},
  journal={Entropy},
  volume={26},
  number={3},
  pages={252},
  year={2024},
  publisher={MDPI}
}

@article{2023multiviewJepa,
  title={Self-supervised information bottleneck for deep multi-view subspace clustering},
  author={Wang, Shiye and Li, Changsheng and Li, Yanming and Yuan, Ye and Wang, Guoren},
  journal={IEEE Transactions on Image Processing},
  volume={32},
  pages={1555--1567},
  year={2023},
  publisher={IEEE}
}

@article{cohen2026horizonImagination,
  title={Horizon Imagination: Efficient On-Policy Rollout in Diffusion World Models},
  author={Cohen, Lior and Nabati, Ofir and Wang, Kaixin and Kumar, Navdeep and Mannor, Shie},
  journal={arXiv preprint arXiv:2602.08032},
  year={2026}
}

@book{joint2008CompositeIndicators,
  title={Handbook on constructing composite indicators: methodology and user guide},
  author={Joint Research Centre},
  year={2008},
  publisher={OECD publishing}
}

@article{how_not_to_lie_with_stats,
  title={How not to lie with statistics: the correct way to summarize benchmark results},
  author={Fleming, Philip J and Wallace, John J},
  journal={Communications of the ACM},
  volume={29},
  number={3},
  pages={218--221},
  year={1986},
  publisher={ACM New York, NY, USA}
}

@article{ding2024diffusion,
  title={Diffusion world model: Future modeling beyond step-by-step rollout for offline reinforcement learning},
  author={Ding, Zihan and Zhang, Amy and Tian, Yuandong and Zheng, Qinqing},
  journal={arXiv preprint arXiv:2402.03570},
  year={2024}
}

@article{jackson2024pgd,
  title={Policy-guided diffusion},
  author={Jackson, Matthew Thomas and Matthews, Michael Tryfan and Lu, Cong and Ellis, Benjamin and Whiteson, Shimon and Foerster, Jakob},
  journal={arXiv preprint arXiv:2404.06356},
  year={2024}
}

@article{lu2023synther,
  title={Synthetic experience replay},
  author={Lu, Cong and Ball, Philip and Teh, Yee Whye and Parker-Holder, Jack},
  journal={Advances in Neural Information Processing Systems},
  volume={36},
  pages={46323--46344},
  year={2023}
}

@inproceedings{van1979information,
  title={Information retrieval: theory and practice},
  author={Van Rijsbergen, C},
  booktitle={Proceedings of the joint IBM/University of Newcastle upon tyne seminar on data base systems},
  volume={79},
  pages={1--14},
  year={1979},
  organization={Butterworth-Heinemann Oxford, UK}
}

@inproceedings{agarwal2021precipice,
  title={Deep reinforcement learning at the edge of the statistical precipice},
  author={Agarwal, Rishabh and Schwarzer, Max and Castro, Pablo Samuel and Courville, Aaron C and Bellemare, Marc},
  booktitle={Advances in Neural Information Processing Systems},
  volume={34},
  pages={29304--29320},
  year={2021}
}

@inproceedings{ronneberger2015UNET,
  title={U-net: Convolutional networks for biomedical image segmentation},
  author={Ronneberger, Olaf and Fischer, Philipp and Brox, Thomas},
  booktitle={Medical image computing and computer-assisted intervention--MICCAI 2015: 18th international conference, Munich, Germany, October 5-9, 2015, proceedings, part III 18},
  pages={234--241},
  year={2015},
  organization={Springer}
}

@article{williams1992REINFORCE,
  title={Simple statistical gradient-following algorithms for connectionist reinforcement learning},
  author={Williams, Ronald J},
  journal={Machine learning},
  volume={8},
  number={3},
  pages={229--256},
  year={1992},
  publisher={Springer}
}

@inproceedings{xiang2023DiffAEAreUnified,
  title={Denoising diffusion autoencoders are unified self-supervised learners},
  author={Xiang, Weilai and Yang, Hongyu and Huang, Di and Wang, Yunhong},
  booktitle={Proceedings of the IEEE/CVF International Conference on Computer Vision},
  pages={15802--15812},
  year={2023}
}

@article{baranchuk2021LabelEfficientUNETFeat,
  title={Label-efficient semantic segmentation with diffusion models},
  author={Baranchuk, Dmitry and Rubachev, Ivan and Voynov, Andrey and Khrulkov, Valentin and Babenko, Artem},
  journal={arXiv preprint arXiv:2112.03126},
  year={2021}
}

@article{nielsen2023diffenc,
  title={Diffenc: Variational diffusion with a learned encoder},
  author={Nielsen, Beatrix MG and Christensen, Anders and Dittadi, Andrea and Winther, Ole},
  journal={arXiv preprint arXiv:2310.19789},
  year={2023}
}

@article{pandey2022diffusevae,
  title={Diffusevae: Efficient, controllable and high-fidelity generation from low-dimensional latents},
  author={Pandey, Kushagra and Mukherjee, Avideep and Rai, Piyush and Kumar, Abhishek},
  journal={arXiv preprint arXiv:2201.00308},
  year={2022}
}

@article{zhang2022UnsupRepLearnPretrainedSEA,
  title={Unsupervised representation learning from pre-trained diffusion probabilistic models},
  author={Zhang, Zijian and Zhao, Zhou and Lin, Zhijie},
  journal={Advances in neural information processing systems},
  volume={35},
  pages={22117--22130},
  year={2022}
}

@article{yue2024ExpDiffTimeSteps,
  title={Exploring diffusion time-steps for unsupervised representation learning},
  author={Yue, Zhongqi and Wang, Jiankun and Sun, Qianru and Ji, Lei and Chang, Eric I and Zhang, Hanwang and others},
  journal={arXiv preprint arXiv:2401.11430},
  year={2024}
}

@inproceedings{yang2023NUSDiffasRepLearner,
  title={Diffusion model as representation learner},
  author={Yang, Xingyi and Wang, Xinchao},
  booktitle={Proceedings of the IEEE/CVF International Conference on Computer Vision},
  pages={18938--18949},
  year={2023}
}

@article{luo2023DiffHyperfeatures,
  title={Diffusion hyperfeatures: Searching through time and space for semantic correspondence},
  author={Luo, Grace and Dunlap, Lisa and Park, Dong Huk and Holynski, Aleksander and Darrell, Trevor},
  journal={Advances in Neural Information Processing Systems},
  volume={36},
  pages={47500--47510},
  year={2023}
}

@article{wang2022DiffPolicyExpressive,
  title={Diffusion policies as an expressive policy class for offline reinforcement learning},
  author={Wang, Zhendong and Hunt, Jonathan J and Zhou, Mingyuan},
  journal={arXiv preprint arXiv:2208.06193},
  year={2022}
}

@article{muzero2020,
  title={Mastering atari, go, chess and shogi by planning with a learned model},
  author={Schrittwieser, Julian and Antonoglou, Ioannis and Hubert, Thomas and Simonyan, Karen and Sifre, Laurent and Schmitt, Simon and Guez, Arthur and Lockhart, Edward and Hassabis, Demis and Graepel, Thore and others},
  journal={Nature},
  volume={588},
  number={7839},
  pages={604--609},
  year={2020},
  publisher={Nature Publishing Group UK London}
}

@article{EfficientZero,
  title={Mastering atari games with limited data},
  author={Ye, Weirui and Liu, Shaohuai and Kurutach, Thanard and Abbeel, Pieter and Gao, Yang},
  journal={Advances in neural information processing systems},
  volume={34},
  pages={25476--25488},
  year={2021}
}

@article{silver2017MuZero,
  title={Mastering chess and shogi by self-play with a general reinforcement learning algorithm},
  author={Silver, David and Hubert, Thomas and Schrittwieser, Julian and Antonoglou, Ioannis and Lai, Matthew and Guez, Arthur and Lanctot, Marc and Sifre, Laurent and Kumaran, Dharshan and Graepel, Thore and others},
  journal={arXiv preprint arXiv:1712.01815},
  year={2017}
}

@article{zhang2023storm,
  title={Storm: Efficient stochastic transformer based world models for reinforcement learning},
  author={Zhang, Weipu and Wang, Gang and Sun, Jian and Yuan, Yetian and Huang, Gao},
  journal={Advances in Neural Information Processing Systems},
  volume={36},
  pages={27147--27166},
  year={2023}
}

@article{micheli2022iris,
  title={Transformers are sample-efficient world models},
  author={Micheli, Vincent and Alonso, Eloi and Fleuret, Fran{\c{c}}ois},
  journal={arXiv preprint arXiv:2209.00588},
  year={2022}
}

@inproceedings{caron2021DINO,
  title={Emerging properties in self-supervised vision transformers},
  author={Caron, Mathilde and Touvron, Hugo and Misra, Ishan and J{\'e}gou, Herv{\'e} and Mairal, Julien and Bojanowski, Piotr and Joulin, Armand},
  booktitle={Proceedings of the IEEE/CVF international conference on computer vision},
  pages={9650--9660},
  year={2021}
}

@article{robine2023TWM,
  title={Transformer-based world models are happy with 100k interactions},
  author={Robine, Jan and H{\"o}ftmann, Marc and Uelwer, Tobias and Harmeling, Stefan},
  journal={arXiv preprint arXiv:2303.07109},
  year={2023}
}

@article{sutton1991DYNA,
  title={Dyna, an integrated architecture for learning, planning, and reacting},
  author={Sutton, Richard S},
  journal={ACM Sigart Bulletin},
  volume={2},
  number={4},
  pages={160--163},
  year={1991},
  publisher={ACM New York, NY, USA}
}

@article{karras2022EDM,
  title={Elucidating the design space of diffusion-based generative models},
  author={Karras, Tero and Aittala, Miika and Aila, Timo and Laine, Samuli},
  journal={Advances in neural information processing systems},
  volume={35},
  pages={26565--26577},
  year={2022}
}

@article{kaiser2019Atari100kSimPLE,
  title={Model-based reinforcement learning for atari},
  author={Kaiser, Lukasz and Babaeizadeh, Mohammad and Milos, Piotr and Osinski, Blazej and Campbell, Roy H and Czechowski, Konrad and Erhan, Dumitru and Finn, Chelsea and Kozakowski, Piotr and Levine, Sergey and others},
  journal={arXiv preprint arXiv:1903.00374},
  year={2019}
}

@article{parkerholder2024genie2,
  title={Genie 2: A large-scale foundation world model},
  author={Parker-Holder, J and Ball, P and Bruce, J and Dasagi, V and Holsheimer, K and Kaplanis, C and Moufarek, A and Scully, G and Shar, J and Shi, J and others},
  journal={URL: https://deepmind. google/discover/blog/genie-2-a-large-scale-foundation-world-model},
  year={2024}
}

@article{blattmann2023StableVideoDiff,
  title={Stable video diffusion: Scaling latent video diffusion models to large datasets},
  author={Blattmann, Andreas and Dockhorn, Tim and Kulal, Sumith and Mendelevitch, Daniel and Kilian, Maciej and Lorenz, Dominik and Levi, Yam and English, Zion and Voleti, Vikram and Letts, Adam and others},
  journal={arXiv preprint arXiv:2311.15127},
  year={2023}
}

@inproceedings{bruce2024genie,
  title={Genie: Generative interactive environments},
  author={Bruce, Jake and Dennis, Michael D and Edwards, Ashley and Parker-Holder, Jack and Shi, Yuge and Hughes, Edward and Lai, Matthew and Mavalankar, Aditi and Steigerwald, Richie and Apps, Chris and others},
  booktitle={Forty-first International Conference on Machine Learning},
  year={2024}
}

@article{hansen2022TDMPC,
  title={Temporal difference learning for model predictive control},
  author={Hansen, Nicklas and Wang, Xiaolong and Su, Hao},
  journal={arXiv preprint arXiv:2203.04955},
  year={2022}
}

@article{lecun2022path,
  title={A path towards autonomous machine intelligence version 0.9. 2, 2022-06-27},
  author={LeCun, Yann and others},
  journal={Open Review},
  volume={62},
  number={1},
  pages={1--62},
  year={2022}
}

@article{song2019NCSN,
  title={Generative modeling by estimating gradients of the data distribution},
  author={Song, Yang and Ermon, Stefano},
  journal={Advances in neural information processing systems},
  volume={32},
  year={2019}
}

@article{tishby2000InfoBottleneck,
  title={The information bottleneck method},
  author={Tishby, Naftali and Pereira, Fernando C and Bialek, William},
  journal={arXiv preprint physics/0004057},
  year={2000}
}

@article{vahdat2021LSGM,
  title={Score-based generative modeling in latent space},
  author={Vahdat, Arash and Kreis, Karsten and Kautz, Jan},
  journal={Advances in neural information processing systems},
  volume={34},
  pages={11287--11302},
  year={2021}
}

@article{hansen2023tdmpc2,
  title={Td-mpc2: Scalable, robust world models for continuous control},
  author={Hansen, Nicklas and Su, Hao and Wang, Xiaolong},
  journal={arXiv preprint arXiv:2310.16828},
  year={2023}
}

@article{hafner2020Dreamerv2,
  title={Mastering atari with discrete world models},
  author={Hafner, Danijar and Lillicrap, Timothy and Norouzi, Mohammad and Ba, Jimmy},
  journal={arXiv preprint arXiv:2010.02193},
  year={2020}
}

@inproceedings{hafner2019Planet,
  title={Learning latent dynamics for planning from pixels},
  author={Hafner, Danijar and Lillicrap, Timothy and Fischer, Ian and Villegas, Ruben and Ha, David and Lee, Honglak and Davidson, James},
  booktitle={International conference on machine learning},
  pages={2555--2565},
  year={2019},
  organization={PMLR}
}

@article{ha2018World,
  title={World models},
  author={Ha, David and Schmidhuber, J{\"u}rgen},
  journal={arXiv preprint arXiv:1803.10122},
  volume={2},
  number={3},
  year={2018}
}

@article{Hafner2020Dreamer,
  title={Dream to control: Learning behaviors by latent imagination},
  author={Hafner, Danijar and Lillicrap, Timothy and Ba, Jimmy and Norouzi, Mohammad},
  journal={arXiv preprint arXiv:1912.01603},
  year={2019}
}

@article{hafner2023Dreamerv3,
  title={Mastering diverse domains through world models},
  author={Hafner, Danijar and Pasukonis, Jurgis and Ba, Jimmy and Lillicrap, Timothy},
  journal={arXiv preprint arXiv:2301.04104},
  year={2023}
}

@article{ho2020ddpm,
  title={Denoising diffusion probabilistic models},
  author={Ho, Jonathan and Jain, Ajay and Abbeel, Pieter},
  journal={Advances in neural information processing systems},
  volume={33},
  pages={6840--6851},
  year={2020}
}

@article{alonso2024diamond,
  title={Diffusion for world modeling: Visual details matter in atari},
  author={Alonso, Eloi and Jelley, Adam and Micheli, Vincent and Kanervisto, Anssi and Storkey, Amos J and Pearce, Tim and Fleuret, Fran{\c{c}}ois},
  journal={Advances in Neural Information Processing Systems},
  volume={37},
  pages={58757--58791},
  year={2024}
}

@article{grill2020BYOL,
  title={Bootstrap your own latent-a new approach to self-supervised learning},
  author={Grill, Jean-Bastien and Strub, Florian and Altch{\'e}, Florent and Tallec, Corentin and Richemond, Pierre and Buchatskaya, Elena and Doersch, Carl and Avila Pires, Bernardo and Guo, Zhaohan and Gheshlaghi Azar, Mohammad and others},
  journal={Advances in neural information processing systems},
  volume={33},
  pages={21271--21284},
  year={2020}
}

@inproceedings{assran2023I-JEPA,
  title={Self-supervised learning from images with a joint-embedding predictive architecture},
  author={Assran, Mahmoud and Duval, Quentin and Misra, Ishan and Bojanowski, Piotr and Vincent, Pascal and Rabbat, Michael and LeCun, Yann and Ballas, Nicolas},
  booktitle={Proceedings of the IEEE/CVF conference on computer vision and pattern recognition},
  pages={15619--15629},
  year={2023}
}

@article{janner2022diffuser,
  title={Planning with diffusion for flexible behavior synthesis},
  author={Janner, Michael and Du, Yilun and Tenenbaum, Joshua B and Levine, Sergey},
  journal={arXiv preprint arXiv:2205.09991},
  year={2022}
}

@article{chi2023DiffusionPolicy,
  title={Diffusion policy: Visuomotor policy learning via action diffusion},
  author={Chi, Cheng and Xu, Zhenjia and Feng, Siyuan and Cousineau, Eric and Du, Yilun and Burchfiel, Benjamin and Tedrake, Russ and Song, Shuran},
  journal={The International Journal of Robotics Research},
  volume={44},
  number={10-11},
  pages={1684--1704},
  year={2025},
  publisher={Sage Publications Sage UK: London, England}
}

@inproceedings{rombach2022StableDiffusion,
  title={High-resolution image synthesis with latent diffusion models},
  author={Rombach, Robin and Blattmann, Andreas and Lorenz, Dominik and Esser, Patrick and Ommer, Bj{\"o}rn},
  booktitle={Proceedings of the IEEE/CVF conference on computer vision and pattern recognition},
  pages={10684--10695},
  year={2022}
}

@inproceedings{esser2024StableDiffusion3,
  title={Scaling rectified flow transformers for high-resolution image synthesis},
  author={Esser, Patrick and Kulal, Sumith and Blattmann, Andreas and Entezari, Rahim and M{\"u}ller, Jonas and Saini, Harry and Levi, Yam and Lorenz, Dominik and Sauer, Axel and Boesel, Frederic and others},
  booktitle={Forty-first international conference on machine learning},
  year={2024}
}

@article{liu2024sora,
  title={Sora: A review on background, technology, limitations, and opportunities of large vision models},
  author={Liu, Yixin and Zhang, Kai and Li, Yuan and Yan, Zhiling and Gao, Chujie and Chen, Ruoxi and Yuan, Zhengqing and Huang, Yue and Sun, Hanchi and Gao, Jianfeng and others},
  journal={arXiv preprint arXiv:2402.17177},
  year={2024}
}

@misc{
bardes2024vjepa,
  title={Revisiting feature prediction for learning visual representations from video},
  author={Bardes, Adrien and Garrido, Quentin and Ponce, Jean and Chen, Xinlei and Rabbat, Michael and LeCun, Yann and Assran, Mahmoud and Ballas, Nicolas},
  journal={arXiv preprint arXiv:2404.08471},
  year={2024}
}

@article{TWISTER,
  title={Learning transformer-based world models with contrastive predictive coding},
  author={Burchi, Maxime and Timofte, Radu},
  journal={arXiv preprint arXiv:2503.04416},
  year={2025}
}

@article{OC-STORM,
  title={Objects matter: object-centric world models improve reinforcement learning in visually complex environments},
  author={Zhang, Weipu and Jelley, Adam and McInroe, Trevor and Storkey, Amos},
  journal={arXiv preprint arXiv:2501.16443},
  year={2025}
}

@article{SIMULUS,
  title={Uncovering Untapped Potential in Sample-Efficient World Model Agents},
  author={Cohen, Lior and Wang, Kaixin and Kang, Bingyi and Gadot, Uri and Mannor, Shie},
  journal={arXiv preprint arXiv:2502.11537},
  year={2025}
}

@article{EA-SIMULUS,
  title={From Observations to Events: Event-Aware World Model for Reinforcement Learning},
  author={Peng, Zhao-Han and Li, Shaohui and Li, Zhi and Ruan, Shulan and Liu, Yu and He, You},
  journal={arXiv preprint arXiv:2601.19336},
  year={2026}
}
\bibliographystyle{unsrtnat}


\appendix

\section{Technical appendices and supplementary material}
\label{sec:proof}
\subsection{JEPA as a Variational Information Bottleneck}
\label{app:jepa-ib}

We consider two observed views or two sequential observations, $x_1$ and $x_2$, together with latent representations $z_1$ and $z_2$. The assumed probabilistic graphical model is
\[
x_1 \leftarrow z_1 \rightarrow z_2 \rightarrow x_2 .
\]
Under this model, the joint distribution factorizes as
\begin{equation}
p_\theta(x_1,x_2,z_1,z_2)
=
p(z_1)\,
p_\theta(z_2\mid z_1)\,
p(x_1\mid z_1)\,
p(x_2\mid z_2).
\label{eq:jepa-factorization}
\end{equation}
Here, $p_\theta(z_2\mid z_1)$ plays the role of the JEPA predictor: it predicts the latent representation of the second view from the latent representation of the first view.

We introduce the amortized variational posterior
\begin{equation}
q_\phi(z_1,z_2\mid x_1,x_2)
=
q_\phi(z_1\mid x_1)q_\phi(z_2\mid x_2).
\label{eq:jepa-variational-posterior}
\end{equation}
For notational compactness, for a fixed pair $(x_1,x_2)$, define
\[
q_1(z_1) := q_\phi(z_1\mid x_1),
\qquad
q_2(z_2) := q_\phi(z_2\mid x_2),
\qquad
q_{12}(z_1,z_2) := q_1(z_1)q_2(z_2).
\]

\paragraph{Variational decomposition of the joint likelihood.}
We begin with the marginal log-likelihood of the two views. Since $q_{12}$ is a normalized distribution over $z_1,z_2$, we may write
\begin{align}
\log p_\theta(x_1,x_2)
&=
\int q_{12}(z_1,z_2)
\log p_\theta(x_1,x_2)
\,\mathrm{d}z_1\mathrm{d}z_2
\notag \\
&=
\int q_{12}(z_1,z_2)
\log
\frac{
p_\theta(x_1,x_2,z_1,z_2)
}{
p_\theta(z_1,z_2\mid x_1,x_2)
}
\,\mathrm{d}z_1\mathrm{d}z_2
\notag \\
&=
\E_{q_{12}}
\left[
\log
\frac{
p_\theta(x_1,x_2,z_1,z_2)
}{
q_{12}(z_1,z_2)
}
\right]
+
\KL
\left(
q_{12}(z_1,z_2)
\,\middle\|\,
p_\theta(z_1,z_2\mid x_1,x_2)
\right).
\label{eq:jepa-var-identity}
\end{align}
Define the posterior gap
\begin{equation}
\Delta_\theta(x_1,x_2)
:=
\KL
\left(
q_{12}(z_1,z_2)
\,\middle\|\,
p_\theta(z_1,z_2\mid x_1,x_2)
\right).
\label{eq:jepa-posterior-gap}
\end{equation}
Substituting the factorization in Eq.~\eqref{eq:jepa-factorization} into Eq.~\eqref{eq:jepa-var-identity}, we obtain
\begin{align}
\log p_\theta(x_1,x_2)
&=
\E_{q_1}
\left[
\log p(x_1\mid z_1)
\right]
+
\E_{q_2}
\left[
\log p(x_2\mid z_2)
\right]
\notag \\
&\quad
-
\E_{q_1}
\left[
\KL
\left(
q_2(z_2)
\,\middle\|\,
p_\theta(z_2\mid z_1)
\right)
\right]
-
\KL
\left(
q_1(z_1)
\,\middle\|\,
p(z_1)
\right)
+
\Delta_\theta(x_1,x_2).
\label{eq:jepa-elbo-decomposition}
\end{align}
The first two terms are view-likelihood terms. The third term is the JEPA prediction term. The fourth term is a bottleneck regularizer on $z_1$, and the final term is the posterior approximation gap.

\paragraph{Connection to mutual information.}
The mutual information between the two views is
\begin{equation}
I(X_1;X_2)
=
\E_{p(x_1,x_2)}
\left[
\log p(x_1,x_2)
-
\log p(x_1)
-
\log p(x_2)
\right].
\label{eq:data-mi}
\end{equation}
Assuming the model marginal matches the population distribution, define the variational estimates
\begin{align}
\widehat I(X_1;Z_1)
&:=
\E_{p(x_1)q_\phi(z_1\mid x_1)}
\left[
\log p(x_1\mid z_1)
-
\log p(x_1)
\right],
\label{eq:ihat-x1z1}
\\
\widehat I(X_2;Z_2)
&:=
\E_{p(x_2)q_\phi(z_2\mid x_2)}
\left[
\log p(x_2\mid z_2)
-
\log p(x_2)
\right].
\label{eq:ihat-x2z2}
\end{align}
Now define the JEPA prediction loss
\begin{equation}
\mathcal L_{\mathrm{JEPA}}
:=
\E_{p(x_1,x_2)q_\phi(z_1\mid x_1)}
\left[
\KL
\left(
q_\phi(z_2\mid x_2)
\,\middle\|\,
p_\theta(z_2\mid z_1)
\right)
\right],
\label{eq:jepa-loss}
\end{equation}
the bottleneck regularizer
\begin{equation}
\mathcal R_1
:=
\E_{p(x_1)}
\left[
\KL
\left(
q_\phi(z_1\mid x_1)
\,\middle\|\,
p(z_1)
\right)
\right],
\label{eq:jepa-bottleneck}
\end{equation}
and the expected posterior gap
\begin{equation}
\mathcal G
:=
\E_{p(x_1,x_2)}
\left[
\KL
\left(
q_\phi(z_1,z_2\mid x_1,x_2)
\,\middle\|\,
p_\theta(z_1,z_2\mid x_1,x_2)
\right)
\right].
\label{eq:jepa-gap}
\end{equation}
Then
\begin{equation}
I(X_1;X_2)
=
\widehat I(X_1;Z_1)
+
\widehat I(X_2;Z_2)
-
\mathcal L_{\mathrm{JEPA}}
-
\mathcal R_1
+
\mathcal G.
\label{eq:jepa-mi-decomposition}
\end{equation}
Equivalently,
\begin{equation}
-\mathcal L_{\mathrm{JEPA}}
=
I(X_1;X_2)
-
\widehat I(X_1;Z_1)
-
\widehat I(X_2;Z_2)
+
\mathcal R_1
-
\mathcal G.
\label{eq:jepa-negative-loss}
\end{equation}
Under the generative Markov chain $X_1 - Z_1 - Z_2 - X_2$, the data processing inequality gives
\begin{equation}
I(X_1;X_2)
\le
I(Z_1;Z_2),
\label{eq:jepa-dpi}
\end{equation}
and therefore
\begin{equation}
-\mathcal L_{\mathrm{JEPA}}
\le
I(Z_1;Z_2)
-
\widehat I(X_1;Z_1)
-
\widehat I(X_2;Z_2)
+
\mathcal R_1
-
\mathcal G.
\label{eq:jepa-ib-bound}
\end{equation}
\subsection{Conditional Diffusion-JEPA as a Variational Information Bottleneck}
\label{app:conditional-diffusion-jepa-ib}

We now extend the variational JEPA argument to a latent conditional diffusion
predictor. The key difference from the one-step JEPA model is that the direct
predictor $p_\theta(z_2\mid z_1)$ is replaced by a conditional reverse diffusion
trajectory that predicts the clean target latent $z_2^0$ from noise while
conditioning on the clean context latent $z_1^0$.

We consider two observed views $x_1$ and $x_2$. The clean context latent is
$z_1^0$, and the target latent is represented by a diffusion path
\[
z_2^{0:T}:=(z_2^0,z_2^1,\ldots,z_2^T),
\]
where $z_2^0$ is the clean target latent and $z_2^T$ is the maximally noised
latent. The assumed graphical model is shown in \cref{fig:joint-embedding-diffusion-pgm},
where each reverse denoising transition is conditioned on $z_1^0$. Equivalently,
the joint distribution factorizes as
\begin{align}
p_\psi(x_1,x_2,z_1^0,z_2^{0:T})
&=
p(z_1^0)\,
p(x_1\mid z_1^0)\,
p(z_2^T)
\prod_{t=1}^{T}
p_\psi(z_2^{t-1}\mid z_2^t,z_1^0)\,
p(x_2\mid z_2^0).
\label{eq:cdj-factorization-simple}
\end{align}
Here,
\[
p_\psi(z_2^{0:T}\mid z_1^0)
:=
p(z_2^T)
\prod_{t=1}^{T}
p_\psi(z_2^{t-1}\mid z_2^t,z_1^0)
\]
is the conditional reverse diffusion path. Therefore, the induced stochastic
JEPA predictor from $z_1^0$ to $z_2^0$ is
\begin{equation}
p_\psi(z_2^0\mid z_1^0)
=
\int
p(z_2^T)
\prod_{t=1}^{T}
p_\psi(z_2^{t-1}\mid z_2^t,z_1^0)
\,\mathrm{d}z_2^{1:T}.
\label{eq:cdj-implicit-predictor-simple}
\end{equation}
Thus, the conditional diffusion model is a multi-step stochastic JEPA predictor.

We introduce the amortized variational posterior
\begin{align}
q_\varphi(z_1^0,z_2^{0:T}\mid x_1,x_2)
&=
q_\varphi(z_1^0\mid x_1)\,
q_\varphi(z_2^0\mid x_2)\,
q(z_2^{1:T}\mid z_2^0),
\label{eq:cdj-variational-posterior-simple}
\end{align}
where $q(z_2^{1:T}\mid z_2^0)$ is the fixed forward diffusion process. For
notational compactness, for a fixed pair $(x_1,x_2)$, define
\[
q_1(z_1^0):=q_\varphi(z_1^0\mid x_1),
\]
\[
q_2^0(z_2^0):=q_\varphi(z_2^0\mid x_2),
\]
\[
q_2^{0:T}(z_2^{0:T})
:=
q_2^0(z_2^0)q(z_2^{1:T}\mid z_2^0),
\]
and
\[
q_{12}(z_1^0,z_2^{0:T})
:=
q_1(z_1^0)q_2^{0:T}(z_2^{0:T}).
\]

\paragraph{Variational decomposition of the joint likelihood.}
We begin with the marginal log-likelihood of the two views. Since $q_{12}$ is a
normalized distribution over $z_1^0,z_2^{0:T}$, we may write
\begin{align}
\log p_\psi(x_1,x_2)
&=
\int q_{12}(z_1^0,z_2^{0:T})
\log p_\psi(x_1,x_2)
\,\mathrm{d}z_1^0\mathrm{d}z_2^{0:T}
\notag \\
&=
\int q_{12}(z_1^0,z_2^{0:T})
\log
\frac{
p_\psi(x_1,x_2,z_1^0,z_2^{0:T})
}{
p_\psi(z_1^0,z_2^{0:T}\mid x_1,x_2)
}
\,\mathrm{d}z_1^0\mathrm{d}z_2^{0:T}
\notag \\
&=
\E_{q_{12}}
\left[
\log
\frac{
p_\psi(x_1,x_2,z_1^0,z_2^{0:T})
}{
q_{12}(z_1^0,z_2^{0:T})
}
\right]
+
\KL
\left(
q_{12}(z_1^0,z_2^{0:T})
\,\middle\|\,
p_\psi(z_1^0,z_2^{0:T}\mid x_1,x_2)
\right).
\label{eq:cdj-var-identity-simple}
\end{align}
Define the posterior gap
\begin{equation}
\Delta_\psi(x_1,x_2)
:=
\KL
\left(
q_{12}(z_1^0,z_2^{0:T})
\,\middle\|\,
p_\psi(z_1^0,z_2^{0:T}\mid x_1,x_2)
\right).
\label{eq:cdj-posterior-gap-simple}
\end{equation}
Substituting the factorization in Eq.~\eqref{eq:cdj-factorization-simple} into
Eq.~\eqref{eq:cdj-var-identity-simple}, we obtain
\begin{align}
\log p_\psi(x_1,x_2)
&=
\E_{q_{12}}
\Bigg[
\log p(x_1\mid z_1^0)
+
\log p(x_2\mid z_2^0)
+
\log p(z_1^0)
\nonumber \\
&\qquad\qquad
+
\log p(z_2^T)
+
\sum_{t=1}^{T}
\log p_\psi(z_2^{t-1}\mid z_2^t,z_1^0)
-
\log q_1(z_1^0)
-
\log q_2^{0:T}(z_2^{0:T})
\Bigg]
\nonumber \\
&\quad
+
\Delta_\psi(x_1,x_2).
\label{eq:cdj-expanded-elbo-raw}
\end{align}
Now group the terms according to their roles. The view-likelihood terms are
\[
\E_{q_{12}}[\log p(x_1\mid z_1^0)]
=
\E_{q_1}[\log p(x_1\mid z_1^0)]
\]
and
\[
\E_{q_{12}}[\log p(x_2\mid z_2^0)]
=
\E_{q_2^0}[\log p(x_2\mid z_2^0)].
\]
The context-prior terms satisfy
\begin{align}
\E_{q_{12}}
\left[
\log p(z_1^0)-\log q_1(z_1^0)
\right]
&=
-
\KL
\left(
q_1(z_1^0)
\,\middle\|\,
p(z_1^0)
\right).
\label{eq:cdj-context-kl-instance}
\end{align}
The target-path terms satisfy
\begin{align}
&\E_{q_{12}}
\left[
\log p(z_2^T)
+
\sum_{t=1}^{T}
\log p_\psi(z_2^{t-1}\mid z_2^t,z_1^0)
-
\log q_2^{0:T}(z_2^{0:T})
\right]
\nonumber \\
&\qquad
=
-
\E_{q_1}
\left[
\KL
\left(
q_2^{0:T}(z_2^{0:T})
\,\middle\|\,
p_\psi(z_2^{0:T}\mid z_1^0)
\right)
\right].
\label{eq:cdj-path-kl-instance-expanded}
\end{align}
Therefore,
\begin{align}
\log p_\psi(x_1,x_2)
&=
\E_{q_1}
\left[
\log p(x_1\mid z_1^0)
\right]
+
\E_{q_2^0}
\left[
\log p(x_2\mid z_2^0)
\right]
\nonumber \\
&\quad
-
\E_{q_1}
\left[
\KL
\left(
q_2^{0:T}(z_2^{0:T})
\,\middle\|\,
p_\psi(z_2^{0:T}\mid z_1^0)
\right)
\right]
\nonumber \\
&\quad
-
\KL
\left(
q_1(z_1^0)
\,\middle\|\,
p(z_1^0)
\right)
+
\Delta_\psi(x_1,x_2).
\label{eq:cdj-elbo-decomposition-simple}
\end{align}
The first two terms are view-likelihood terms. The third term is the
conditional diffusion-JEPA prediction term. The fourth term is a bottleneck
regularizer on the clean context latent $z_1^0$, and the final term is the
posterior approximation gap.

Define the instance-level conditional diffusion path loss
\begin{equation}
\mathcal L_{\mathrm{CDJ}}^{\mathrm{path}}(x_1,x_2)
:=
\E_{q_1}
\left[
\KL
\left(
q_2^{0:T}(z_2^{0:T})
\,\middle\|\,
p_\psi(z_2^{0:T}\mid z_1^0)
\right)
\right].
\label{eq:cdj-path-loss-instance-simple}
\end{equation}
This is the direct analogue of the one-step JEPA prediction loss
$\E_{q_1}[\KL(q_2(z_2)\|p_\theta(z_2\mid z_1))]$, except that the predicted
object is now the entire reverse diffusion path $z_2^{0:T}$.

To see the diffusion terms explicitly, write
\begin{align}
\mathcal L_{\mathrm{CDJ}}^{\mathrm{path}}(x_1,x_2)
&=
\E_{q_1 q_2^{0:T}}
\Bigg[
\log q_2^{0:T}(z_2^{0:T})
-
\log p(z_2^T)
-
\sum_{t=1}^{T}
\log p_\psi(z_2^{t-1}\mid z_2^t,z_1^0)
\Bigg].
\label{eq:cdj-path-loss-log-ratio}
\end{align}
Since
\[
q_2^{0:T}(z_2^{0:T})
=
q_2^0(z_2^0)q(z_2^{1:T}\mid z_2^0),
\]
we may also write
\begin{align}
\mathcal L_{\mathrm{CDJ}}^{\mathrm{path}}(x_1,x_2)
&=
\E_{q_1 q_2^0 q(z_2^{1:T}\mid z_2^0)}
\Bigg[
\log q_2^0(z_2^0)
+
\log q(z_2^{1:T}\mid z_2^0)
\nonumber \\
&\qquad\qquad
-
\log p(z_2^T)
-
\sum_{t=1}^{T}
\log p_\psi(z_2^{t-1}\mid z_2^t,z_1^0)
\Bigg].
\label{eq:cdj-path-loss-forward-expanded}
\end{align}
For a Markov forward diffusion process, this path loss admits the usual DDPM
decomposition. Using
\[
q(z_2^{1:T}\mid z_2^0)
=
q(z_2^T\mid z_2^0)
\prod_{t=2}^{T}
q(z_2^{t-1}\mid z_2^t,z_2^0),
\]
we obtain
\begin{align}
\mathcal L_{\mathrm{CDJ}}^{\mathrm{path}}(x_1,x_2)
&=
-H(q_2^0)
+
\E_{q_2^0}
\left[
\KL
\left(
q(z_2^T\mid z_2^0)
\,\middle\|\,
p(z_2^T)
\right)
\right]
\nonumber \\
&\quad
+
\E_{q_1 q_2^0 q(z_2^1\mid z_2^0)}
\left[
-\log p_\psi(z_2^0\mid z_2^1,z_1^0)
\right]
\nonumber \\
&\quad
+
\sum_{t=2}^{T}
\E_{q_1 q_2^0 q(z_2^t\mid z_2^0)}
\left[
\KL
\left(
q(z_2^{t-1}\mid z_2^t,z_2^0)
\,\middle\|\,
p_\psi(z_2^{t-1}\mid z_2^t,z_1^0)
\right)
\right].
\label{eq:cdj-path-loss-ddpm-expanded}
\end{align}
The first term is the entropy of the target encoder distribution, the second
term is the terminal prior-matching term, the third term is the endpoint
reconstruction term for $z_2^0$, and the remaining terms are the conditional
denoising KLs.

\paragraph{Connection to mutual information.}
The mutual information between the two views is
\begin{equation}
I(X_1;X_2)
=
\E_{p(x_1,x_2)}
\left[
\log p(x_1,x_2)
-
\log p(x_1)
-
\log p(x_2)
\right].
\label{eq:cdj-data-mi-simple}
\end{equation}
Assuming the model marginal matches the population distribution, define the
variational estimates
\begin{align}
\widehat I(X_1;Z_1^0)
&:=
\E_{p(x_1)q_\varphi(z_1^0\mid x_1)}
\left[
\log p(x_1\mid z_1^0)
-
\log p(x_1)
\right],
\label{eq:cdj-ihat-x1z10}
\\
\widehat I(X_2;Z_2^0)
&:=
\E_{p(x_2)q_\varphi(z_2^0\mid x_2)}
\left[
\log p(x_2\mid z_2^0)
-
\log p(x_2)
\right].
\label{eq:cdj-ihat-x2z20}
\end{align}
Now define the population-level conditional diffusion-JEPA path loss
\begin{equation}
\mathcal L_{\mathrm{CDJ}}^{\mathrm{path}}
:=
\E_{p(x_1,x_2)}
\left[
\mathcal L_{\mathrm{CDJ}}^{\mathrm{path}}(x_1,x_2)
\right],
\label{eq:cdj-path-loss-pop-simple}
\end{equation}
the bottleneck regularizer
\begin{equation}
\mathcal R_1
:=
\E_{p(x_1)}
\left[
\KL
\left(
q_\varphi(z_1^0\mid x_1)
\,\middle\|\,
p(z_1^0)
\right)
\right],
\label{eq:cdj-bottleneck-simple}
\end{equation}
and the expected posterior gap
\begin{equation}
\mathcal G
:=
\E_{p(x_1,x_2)}
\left[
\KL
\left(
q_\varphi(z_1^0,z_2^{0:T}\mid x_1,x_2)
\,\middle\|\,
p_\psi(z_1^0,z_2^{0:T}\mid x_1,x_2)
\right)
\right].
\label{eq:cdj-gap-simple}
\end{equation}
Taking the expectation of Eq.~\eqref{eq:cdj-elbo-decomposition-simple} over
$p(x_1,x_2)$ and subtracting $\log p(x_1)+\log p(x_2)$ gives
\begin{equation}
I(X_1;X_2)
=
\widehat I(X_1;Z_1^0)
+
\widehat I(X_2;Z_2^0)
-
\mathcal L_{\mathrm{CDJ}}^{\mathrm{path}}
-
\mathcal R_1
+
\mathcal G.
\label{eq:cdj-mi-decomposition-simple}
\end{equation}
Equivalently,
\begin{equation}
-\mathcal L_{\mathrm{CDJ}}^{\mathrm{path}}
=
I(X_1;X_2)
-
\widehat I(X_1;Z_1^0)
-
\widehat I(X_2;Z_2^0)
+
\mathcal R_1
-
\mathcal G.
\label{eq:cdj-negative-loss-simple}
\end{equation}

After marginalizing the intermediate diffusion variables $z_2^{1:T}$, the model
induces the clean-latent Markov structure
\[
X_1 - Z_1^0 - Z_2^0 - X_2.
\]
Therefore, by the data processing inequality,
\begin{equation}
I(X_1;X_2)
\le
I(Z_1^0;Z_2^0).
\label{eq:cdj-dpi-simple}
\end{equation}
Substituting this into Eq.~\eqref{eq:cdj-negative-loss-simple} gives
\begin{equation}
-\mathcal L_{\mathrm{CDJ}}^{\mathrm{path}}
\le
I(Z_1^0;Z_2^0)
-
\widehat I(X_1;Z_1^0)
-
\widehat I(X_2;Z_2^0)
+
\mathcal R_1
-
\mathcal G.
\label{eq:cdj-ib-bound-simple}
\end{equation}

Finally, if one optimizes only the DDPM-style learnable denoising part of the
path objective, define
\begin{align}
\mathcal L^0
&:=
\E_{p(x_1,x_2)}
\E_{q_1 q_2^0 q(z_2^1\mid z_2^0)}
\left[
-\log p_\psi(z_2^0\mid z_2^1,z_1^0)
\right],
\nonumber \\
\mathcal L_{\mathrm{CDJ}}^{\mathrm{den}}
&:=
\mathcal L^0
+
\E_{p(x_1,x_2)}
\sum_{t=2}^{T}
\E_{q_1 q_2^0 q(z_2^t\mid z_2^0)}
\left[
\KL
\left(
q(z_2^{t-1}\mid z_2^t,z_2^0)
\,\middle\|\,
p_\psi(z_2^{t-1}\mid z_2^t,z_1^0)
\right)
\right]
.
\label{eq:cdj-denoising-loss-simple}
\end{align}
Then
\begin{equation}
\mathcal L_{\mathrm{CDJ}}^{\mathrm{path}}
=
\mathcal C_2
+
\mathcal R_{2,T}
+
\mathcal L_{\mathrm{CDJ}}^{\mathrm{den}},
\label{eq:cdj-path-den-decomp-simple}
\end{equation}
where
\begin{equation}
\mathcal R_{2,T}
:=
\E_{p(x_2)q_\varphi(z_2^0\mid x_2)}
\left[
\KL
\left(
q(z_2^T\mid z_2^0)
\,\middle\|\,
p(z_2^T)
\right)
\right]
\label{eq:cdj-terminal-regularizer-simple}
\end{equation}
is the terminal prior-matching term, while
\begin{equation}
\mathcal C_2
:=
\E_{p(x_2)}
\left[
-H\bigl(q_\varphi(z_2^0\mid x_2)\bigr)
\right]
=
\E_{p(x_2)q_\varphi(z_2^0\mid x_2)}
\left[
\log q_\varphi(z_2^0\mid x_2)
\right]
\label{eq:cdj-c2-simple}
\end{equation}
is the $\psi$-independent remainder. Substituting
Eq.~\eqref{eq:cdj-path-den-decomp-simple} into
Eq.~\eqref{eq:cdj-ib-bound-simple} yields
\begin{align}
-\mathcal L_{\mathrm{CDJ}}^{\mathrm{den}}
&\le
I(Z_1^0;Z_2^0)
-
\widehat I(X_1;Z_1^0)
-
\widehat I(X_2;Z_2^0)
\nonumber \\
&\quad
+
\mathcal R_1
+
\mathcal R_{2,T}
-
\mathcal G
+
\mathcal C_2.
\label{eq:cdj-denoising-only-ib-bound-simple}
\end{align}
Therefore, even when training is written only as a conditional denoising
objective, the conditional diffusion-JEPA loss exhibits the same variational
information-bottleneck structure as the one-step JEPA loss, up to terminal
diffusion regularization, endpoint constants, and the variational posterior
gap.

\subsubsection{Remark on optimizing the bottleneck regularizer.}
\label{app:optimizing-the-bottlenck-regularizer}
The decomposition above shows that the exact variational information-bottleneck
objective is
\begin{equation}
\min_{\varphi,\psi}
\mathcal L_{\mathrm{CDJ}}^{\mathrm{path}}
+
\mathcal R_1,
\end{equation}
or, more generally,
\begin{equation}
\min_{\varphi,\psi}
\mathcal L_{\mathrm{CDJ}}^{\mathrm{path}}
+
\beta \mathcal R_1.
\end{equation}
In our experiments, we do not directly optimize the exact KL regularizer
$\mathcal R_1$. Accordingly, the most precise interpretation of JEDI is that it
optimizes the conditional diffusion prediction term that appears inside the
variational information-bottleneck decomposition, rather than exactly
optimizing the full variational information-bottleneck objective. We therefore
present the bottleneck view as a theoretical justification and design lens for
the current method, while treating explicit optimization of $\mathcal R_1$ as a
promising extension.

\paragraph{Relation to VICReg and LeJEPA \citep{balestriero2025lejepa, shwartz2023informationVICREG}}
\label{app:relation-vicereg-lejepa}
The bottleneck regularizer $\mathcal R_1$ also clarifies the connection between
our variational view and recent explicit regularization methods for
collapse-free joint embedding learning. We do not incorporate such
regularizers here, since our empirical study focuses on conventional JEPA
training, but the decomposition reveals a close conceptual link.

By the standard decomposition,
\begin{equation}
\mathcal R_1
=
\E_{p(x_1)}
\KL\left(
q_\varphi(z_1^0\mid x_1)
\,\middle\|\,
p(z_1^0)
\right)
=
I_q(X_1;Z_1^0)
+
\KL\left(
q_\varphi(z_1^0)
\,\middle\|\,
p(z_1^0)
\right).
\end{equation}
The first term is the information-bottleneck component, which explicitly
penalizes the information retained by the context representation. The second
term is an aggregate distribution-matching regularizer. When
$p(z_1^0)=\mathcal N(0,I)$, this aggregate term encourages the representation
distribution to be centered, non-collapsed, and isotropic. This is closely
related in spirit to the variance and covariance regularizers used in VICReg,
which maintain per-dimension variance and decorrelate embedding coordinates, and
to the SIGReg regularizer used in LeJEPA, which directly encourages embeddings
to match an isotropic Gaussian distribution.

Thus, $\mathcal R_1$ can be viewed as a variational counterpart to these
explicit representation regularizers, with an additional
information-bottleneck interpretation.

\subsubsection{Remark on $\mathcal{C}_2$ remainder term}
With the exact DDPM KL decomposition, the remainder term has the explicit form
\begin{equation}
\mathcal C_2(x_2)
:=
-H(q_\varphi(z_2^0\mid x_2))
=
\E_{q_\varphi(z_2^0\mid x_2)}
\left[
\log q_\varphi(z_2^0\mid x_2)
\right].
\end{equation}
At the population level,
\begin{equation}
\mathcal C_2
=
\E_{p(x_2)}
\left[
\E_{q_\varphi(z_2^0\mid x_2)}
\log q_\varphi(z_2^0\mid x_2)
\right].
\end{equation}
Thus,
\begin{equation}
\mathcal L_{\mathrm{CDJ}}^{\mathrm{path}}
=
\mathcal C_2
+
\mathcal R_{2,T}
+
\mathcal L_{\mathrm{CDJ}}^{\mathrm{den}}.
\end{equation}
When $q_\varphi(z_2^0\mid x_2)$ has fixed entropy, for example under a
deterministic or fixed-variance target encoder with stop-gradient, $\mathcal C_2$
is constant with respect to the conditional reverse parameters $\psi$. If the
target encoder distribution is learnable through this objective, however,
$\mathcal C_2$ may depend on $\varphi$ and should not be treated as a global
constant.

\newpage
\subsection{JEDI Algorithm}
Green font refers to changes as compared to pixel-based diffusion world model baseline \citep{alonso2024diamond}
\begin{algorithm}[t]
\scriptsize
\caption{JEDI Training}
\label{alg:jedi}
\begin{algorithmic}[1]
\Procedure{training\_loop}{}
    \For{epochs}
        \State collect\_experience(steps\_collect)
        \For{steps\_diffusion\_model}
            \State \textcolor{darkgreen}{update\_latent\_diffusion\_model()}
        \EndFor
        \For{steps\_reward\_end\_model}
            \State update\_reward\_end\_model()
        \EndFor
        \For{steps\_actor\_critic}
            \State update\_actor\_critic()
        \EndFor
    \EndFor
\EndProcedure
\Statex
\Procedure{collect\_experience}{$n$}
    \State $x_0 \leftarrow \text{env.reset}()$
    \For{$t=0$ \textbf{to} $n-1$}
        \State Sample $a_t \sim \textcolor{darkgreen}{\pi_\omega(a_t | z_t^0)q_\phi(z_t^0|x_t)}$ \Comment{derive $z_t^0$ using JEDI encoder $\mathbf{E}_\phi$ before sampling action}
        \State $x_{t+1}, r_t, d_t \leftarrow \text{env.step}(a_t)$
        \State $\mathcal{D} \leftarrow \mathcal{D} \cup \{x_t, a_t, r_t, d_t\}$
        \If{$d_t = \text{true}$}
            \State $x_{t+1} \leftarrow \text{env.reset}()$
        \EndIf
    \EndFor
\EndProcedure
\Statex
\Procedure{\textcolor{darkgreen}{update\_latent\_diffusion\_model}}{}
    \State Sample sequence $(x_{t-L+1}, a_{t-L+1}, \dots, x_t, a_t, x_{t+1}) \sim \mathcal{D}$
    \State Sample $\log(\sigma) \sim \mathcal{N}(P_{\text{mean}}, P_{\text{std}}^2)$ \Comment{log-normal sigma distribution from EDM}
    \State Define $\tau := \sigma$ \Comment{default identity schedule from EDM}
    \State \textcolor{darkgreen}{compute $[z_{t-L+1}^0, \dots, z_t^0, z_{t+1}^0] = C(\mathbf{E}_\phi([x_{t-L+1}, \dots, x_t, x_{t+1}])$)} \Comment {derive the clamped latents}
    \State \textcolor{darkgreen}{derive $z_{t+1}^0 = \textnormal{sg(}z_{t+1}^0\textnormal{)}$ } \Comment {detach gradients from the target latent}
    \State Sample $\textcolor{darkgreen}{{z}_{t+1}^\tau} \sim \mathcal{N}(\textcolor{pastelgreen}{z_{t+1}^0}, \sigma^2 \mathbf{I})$ \Comment{add independent Gaussian noise}
    \State Compute \textcolor{darkgreen}{$\hat{z}_{t+1}^{0} = \mathbf{D}_\theta({z}_{t+1}^\tau, \tau, z_{k-L+1}^0, a_{t-L+1}, \dots, z_t^0, a_t)$}
    \State Compute reconstruction loss $\mathcal{L}(\theta) = \| \textcolor{darkgreen}{\hat{z}_{t+1}^{0} - z_{t+1}^0} \|^2$
    \State \textcolor{darkgreen}{$rs \;\sim\; \mathrm{Uniform}\{\,\text{True},\text{False}\,\}$} 
    \If{\textcolor{darkgreen}{$rs$ = True}} \Comment{random switch between $\hat{z}_{t+1}^0$ or ${z}_{t+1}^0$ as $\mathbf{D}_\theta$ input for loss at $t+1$}
        \State \textcolor{darkgreen}{Cache $\hat{z}_{t+1}^0$ as input to $\mathbf{D}_\theta$ for {\scriptsize UPDATE\_LATENT\_DIFFUSION\_MODEL} at $t+1$}
    \EndIf
    \State Update $\mathbf{D}_\theta$
\EndProcedure
\Statex
\Procedure{update\_reward\_end\_model}{}
    \State Sample indexes $\mathcal{I} = \{t, \dots, t+L+H-1\}$ \Comment{burn-in + imagination horizon}
    \State Sample sequence $(x_i, a_i, r_i, d_i)_{i \in \mathcal{I}} \sim \mathcal{D}$
    \State \textcolor{darkgreen}{Compute $(z_{i}^0)_{i\in \mathcal{I}} = C(\mathbf{E}_\phi((x_{i})_{i\in \mathcal{I}})$)} \Comment {derive the clamped latents}
    \State Initialize $h=c=0$ \Comment{LSTM hidden and cell states}
    \For{$i \in \mathcal{I}$ \textbf{do}}
        \State Compute $\hat{r}_i, \hat{d}_i, h, c = R_\psi(\textcolor{darkgreen}{z_i^0}, a_i, h, c)$
    \EndFor
    \State Compute $\mathcal{L}(\psi) = \sum_{i \in \mathcal{I}} \text{CE}(\hat{r}_i, \text{sign}(r_i)) + \text{CE}(\hat{d}_i, d_i)$ \Comment{CE: cross-entropy loss}
    \State Update $R_\psi$
\EndProcedure
\Statex
\Procedure{update\_actor\_critic}{}
    \State Sample initial buffer $(x_{t-L+1}, a_{t-L+1}, \dots, x_t) \sim \mathcal{D}$
    \State \textcolor{darkgreen}{Compute $[z_{t-L+1}^0, \dots, z_t^0] = C(\mathbf{E}_\phi([x_{t-L+1}, \dots, x_t])$)} \Comment {derive the clamped latents}
    \State Burn-in buffer with $R_\psi, \pi_\omega$ and $V_\omega$ to initialize LSTM states
    \For{$i=t$ \textbf{to} $t+H-1$}
        \State Sample $a_i \sim \pi_\omega(a_i | \textcolor{darkgreen}{z_i^0})$
        \State Sample reward $r_i$ and termination $d_i$ with $R_\psi$
        \State Sample next observation \textcolor{darkgreen}{$z_{i+1}^0$} by simulating reverse diffusion process with $\mathbf{D}_\theta$
    \EndFor
    \State Compute $V_\omega(\textcolor{darkgreen}{z_t^0})$ for $i = t, \dots, t+H$
    \State Compute RL losses $\mathcal{L}_V(\omega)$ and $\mathcal{L}_\pi(\omega)$
    \State Update $\pi_\omega$ and $V_\omega$
\EndProcedure
\end{algorithmic}
\end{algorithm}

\clearpage
\newpage

\section{Changes Compared to Baseline and Hyperparameters}
\label{appendix:hyperparameters}
We used the DIAMOND library \citep{alonso2024diamond} (MIT License) for our implementation. The main difference between JEDI World Model and DIAMOND is the transplant of the DIAMOND's RL encoder onto the JEDI World Model and the techniques implemented to facilitate end-to-end latent diffusion mentioned in \cref{sec:method_jedi}. The only network architectural changes are practical: (1) changing the original RL encoder to downsample to $z_t \in [-3,3]^{16\times8\times8}$, (2) modifying the diffusion UNET \citep{ronneberger2015UNET} into a single layer without downsampling, (3) and removing the downsampling in the Reward/Termination Model's encoder. The only hyperparameter changes were increasing the diffusion model's warm-up learning steps to 1000 and sigma of input data to be 1. 

\subsection{Benchmark and environment settings}
\label{appendix:benchmark-settings}
\paragraph{Atari100k.}
We follow the standard Atari100k protocol used by prior online world-model work and the IRIS evaluation codebase. In particular, each run uses the 26-game Atari100k benchmark with a total budget of $100$k environment interactions, and we report results over 5 seeds per task as described in \cref{sec:results}. Observations are RGB frames resized to $64\times 64$, matching the image interface used by the JEDI encoder and world model. We use the standard Atari v4 setup with frame skip 4, which matches the protocol used in prior Atari100k world-model evaluations.

\paragraph{Evaluation metrics.}
Following \citet{agarwal2021precipice}, we compute aggregate Atari100k metrics from human-normalized scores (HNS). For a game with agent score $S_{\mathrm{agent}}$, random-policy score $S_{\mathrm{random}}$, and human score $S_{\mathrm{human}}$, the human-normalized score is
\[
\mathrm{HNS}
=
\frac{S_{\mathrm{agent}} - S_{\mathrm{random}}}{S_{\mathrm{human}} - S_{\mathrm{random}}}.
\]
An HNS of $1$ corresponds to human-level performance. ``Mean'' refers to the arithmetic mean of HNS values across the benchmark. ``IQM'' (interquartile mean) is the mean of the middle 50\% of HNS values across game seed runs, which makes it more robust to outliers than the plain mean. ``Optimality Gap'' measures how far performance remains below human level, and is reported as the average gap to the threshold $\mathrm{HNS}=1$, i.e., $\mathbb{E}[\max(1-\mathrm{HNS},0)]$, so lower is better.

\paragraph{Craftium.}
For Craftium, we evaluate on four tasks with 3 seeds per task, as described in \cref{sec:results}. We use standard visual observations with RGB images resized to $64\times 64$, so that the same encoder and latent world-model interface used on Atari can be applied directly in the 3D setting. We use 300 training steps and 200 collection steps per epoch. We report the training returns. We leverage the symlog reward prediction used in HI \citep{cohen2026horizonImagination} for the continuous reward task, Speleo. We also follow HI and use 30k environment steps for SmallRoom-v0 and 100k environment steps for the other 3 environments

\subsection{Third-party assets and licenses}
\label{appendix:asset-licenses}
We explicitly credit the third-party assets used in this work and follow their stated licenses and terms of use. Our implementation builds on the DIAMOND library \citep{alonso2024diamond}, which is released under the MIT License. For Atari100k evaluation, we used the publicly available IRIS codebase associated with \citet{micheli2022iris}, which is released under the GNU General Public License v3.0. For the Craftium experiments, we used the Craftium environment \citep{malagon2024craftium}; its repository states that the project inherits the Luanti/Minetest licensing scheme, with code under the MIT License and game content under CC-BY-SA 3.0. These assets are cited in the bibliography and used in accordance with their repository-stated licenses.

\begin{table}[t]
\centering
\caption{Model architecture details for JEDI. Green font refers to changes as compared to pixel-based diffusion world model baseline \citep{alonso2024diamond}. Downsampling layers refers to the Maxpool operation that downsamples the input height and width by a factor of 2.}
\begin{tabular}{l@{\hspace{2em}}c}
\toprule
\textbf{Hyperparameter} & \textbf{Value} \\
\midrule
\multicolumn{2}{l}{\textbf{Latent Diffusion Dynamics Model ($\mathbf{D}_\theta$)}} \\
State conditioning mechanism & Frame stacking \\
Action conditioning mechanism & Adaptive Group Normalization \\
Diffusion time conditioning mechanism & Adaptive Group Normalization \\
Residual blocks layers & \textcolor{darkgreen}{[1]} \\
Residual blocks channels & \textcolor{darkgreen}{[160]} \\
UNET Downsampling &  \textcolor{darkgreen}{NIL} \\
Residual blocks conditioning dimension & 256 \\
Sigma data (for preconditioning) & \textcolor{darkgreen}{1} \\
\addlinespace
\textcolor{darkgreen}{\textbf{World Model Encoder ($\mathbf{E}_\phi$)}} \\
\textcolor{darkgreen}{{Encoder blocks layers}} & \textcolor{darkgreen}{[1, 1, 1, 1]} \\
\textcolor{darkgreen}{{Encoder blocks channels}} & \textcolor{darkgreen}{[32, 32, 32, 16]} \\
\textcolor{darkgreen}{Encoder Downsampling layers} &  \textcolor{darkgreen}{[1,1,1,0]} \\
\textcolor{darkgreen}{Encoder tanh clamp factor} & \textcolor{darkgreen}{3} \\
\addlinespace

\multicolumn{2}{l}{\textbf{Reward/Termination Model ($\mathbf{R}_\psi$)}} \\
Action conditioning mechanisms & Adaptive Group Normalization \\
Residual blocks layers & {[2, 2, 2, 2]} \\
Residual blocks channels & {[32, 32, 32, 32]} \\
Encoder Downsampling layers &  \textcolor{darkgreen}{NIL} \\
Residual blocks conditioning dimension & 128 \\
LSTM dimension & 512 \\
\addlinespace

\multicolumn{2}{l}{\textbf{Actor-Critic Model ($\pi_\omega$ and $V_\omega$)}} \\
Encoder & \textcolor{darkgreen}{NIL} \\
LSTM dimension & 512 \\
\bottomrule
\end{tabular}
\end{table}

\begin{table}[t]
\centering
\caption{Training hyperparameters for JEDI. Green font refers to changes as compared to pixel-based diffusion world model baseline \citep{alonso2024diamond}.}
\resizebox{\linewidth}{!}{%
\begin{tabular}{l@{\hspace{2em}}c}
\toprule
\textbf{Hyperparameter} & \textbf{Value} \\
\midrule

\multicolumn{2}{l}{\textbf{Training loop}} \\
Number of epochs & 1000 \\
Training steps per epoch & 400 \\
Denoiser Batch size & \textcolor{pastelgreen}{32/64} \\
Other Batch size & {32} \\

Environment steps per epoch & 100 \\
Epsilon (greedy) for collection & 0.01 \\
\addlinespace

\multicolumn{2}{l}{\textbf{RL hyperparameters}} \\
Imagination horizon ($H$) & 15 \\
Discount factor ($\gamma$) & 0.985 \\
Entropy weight ($\eta$) & 0.001 \\
$\lambda$-returns coefficient ($\lambda$) & 0.95 \\
\addlinespace

\multicolumn{2}{l}{\textbf{Sequence construction during training}} \\
For $D_\theta$, number of conditioning observations and actions ($L$) & 4 \\
For $R_\psi$, burn-in length ($B_R$), set to $L$ in practice & 4 \\
For $R_\psi$, training sequence length ($B_R + H$) & 19 \\
For $\pi_\phi$ and $V_\phi$, burn-in length ($B_{\pi,V}$), set to $L$ in practice & 4 \\
\addlinespace

\multicolumn{2}{l}{\textbf{Optimization}} \\
Optimizer & AdamW \\
Learning rate & 1e-4 \\
Epsilon & 1e-8 \\
Weight decay ($D_\theta$) & 1e-2 \\
Weight decay ($R_\psi$) & 1e-2 \\
Weight decay ($\pi_\phi$ and $V_\phi$) & 0 \\
Learning rate Warm-up steps ($\mathbf{D}_\theta$) & \textcolor{darkgreen}{1e3} \\
Learning rate Warm-up steps($\mathbf{R}_\psi$) & 1e2 \\
Learning rate Warm-up steps ($\pi_\omega$ and $V_\omega$) & 1e2 \\
\textcolor{darkgreen}{Learning rate scale factor for $\mathbf{E_\phi}$} & \textcolor{darkgreen}{0.3} \\
\addlinespace

\multicolumn{2}{l}{\textbf{Diffusion Sampling}} \\
Method & Euler \\
Number of steps & 3 \\
S churn & \textcolor{darkgreen}{1 (only for stochastic experiments)}\\
\addlinespace

\multicolumn{2}{l}{\textbf{Environment}} \\
Image observation dimensions & $64 \times 64 \times 3$ \\
Action space & Discrete (up to 18 actions) \\
Frameskip & 4 \\
Frameskip for Stochastic Experiments & [2, 6] \\
Max noop & 30 \\
Termination on life loss & True \\
Reward clipping & \{-1, 0, 1\} \\
\bottomrule
\end{tabular}%
}
\end{table}

\newpage
\pagebreak
\clearpage

\section{Action Space and Shooter Tasks}
\newcommand{\cmark}{\textcolor{green!60!black}{\ding{51}}} 
\newcommand{\xmark}{\textcolor{red}{\ding{55}}} 

\begin{table}[h]  
\renewcommand{\arraystretch}{0.8}
\centering
\caption{Atari100k tasks with number of actions and shooter-task label.}
\label{tab:tasks_with_num_actions_shooter_task}
\resizebox{0.55\linewidth}{!}{%
\begin{tabular}{|l|c|c|}
\toprule
Task                      & Num Actions   & Shooter Task? \\
\midrule
\textit{Boxing}           & 18            & \xmark         \\
\textit{Krull}            & 18            & \xmark         \\
\textit{CrazyClimber}     & 9             & \xmark         \\
\textit{Gopher}           & 8             & \xmark         \\
\textit{RoadRunner}       & 18            & \xmark         \\
\textit{Jamesbond}        & 18            & \cmark         \\
\textit{Assault}          & 7             & \cmark         \\
\textit{Breakout}         & 4             & \xmark         \\
\textit{KungFuMaster}     & 14            & \xmark         \\
\textit{Pong}             & 6             & \xmark         \\
\textit{Kangaroo}         & 18            & \xmark         \\
\textit{UpNDown}          & 6             & \xmark         \\
\textit{Freeway}          & 3             & \xmark         \\
\textit{BankHeist}        & 18            & \cmark         \\
\textit{DemonAttack}      & 6             & \cmark         \\
\textit{Hero}             & 18            & \cmark         \\
\textit{BattleZone}       & 18            & \cmark         \\
\textit{Frostbite}        & 18            & \xmark         \\
\textit{Qbert}            & 6             & \xmark         \\
\textit{MsPacman}         & 9             & \xmark         \\
\textit{Asterix}          & 9             & \xmark         \\
\textit{ChopperCommand}   & 18            & \cmark         \\
\textit{Amidar}           & 10            & \xmark         \\
\textit{Alien}            & 18            & \cmark         \\
\textit{PrivateEye}       & 18            & \xmark         \\
\textit{Seaquest}         & 18            & \cmark         \\
\bottomrule
\end{tabular}
}
\renewcommand{\arraystretch}{1}
\end{table}

\section{Supplementary Results}

\begin{figure*}[t]
    \centering
    \includegraphics[width=0.95\textwidth]{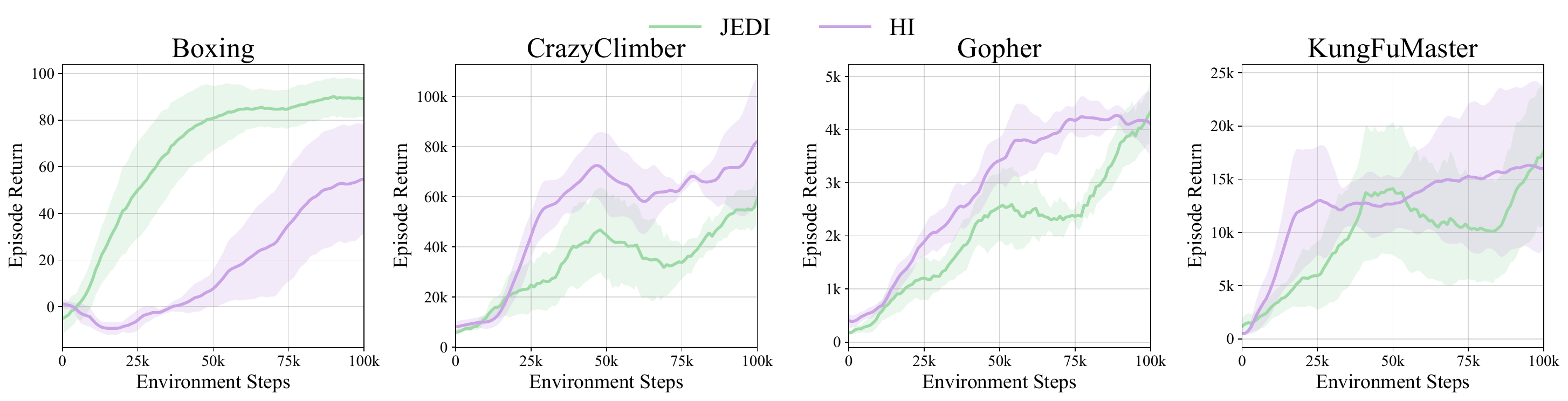}
    \caption{JEDI versus HI training curves on Atari100k}
    \label{fig:JEDI_vs_HI_atari_curves}
\end{figure*}

\begin{figure}[t]
  \centering
  \begin{subfigure}[t]{0.25\textwidth}
    \centering
    \includegraphics[width=\textwidth]{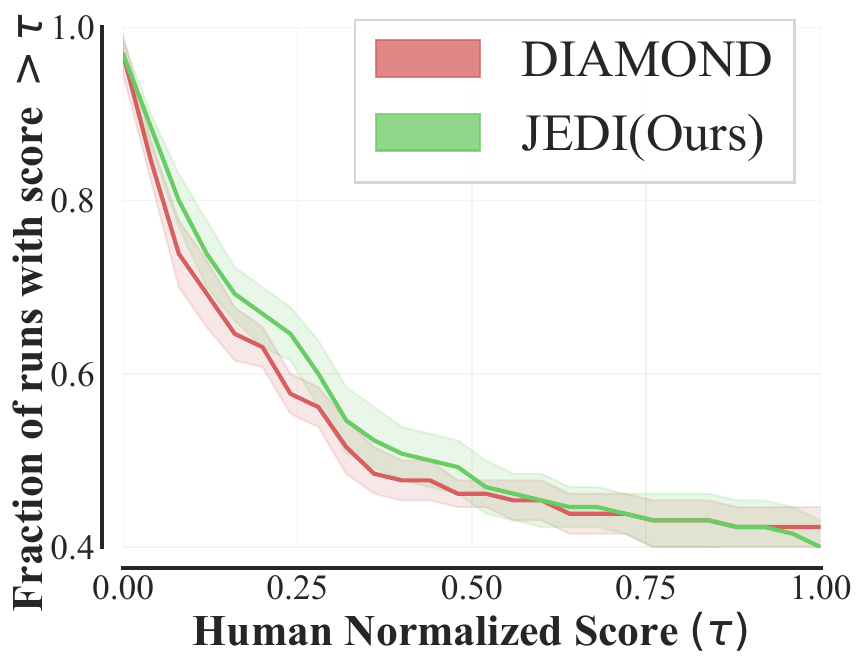}
  \end{subfigure}
  \begin{subfigure}[t]{0.25\textwidth}
    \centering
    \includegraphics[width=\textwidth]{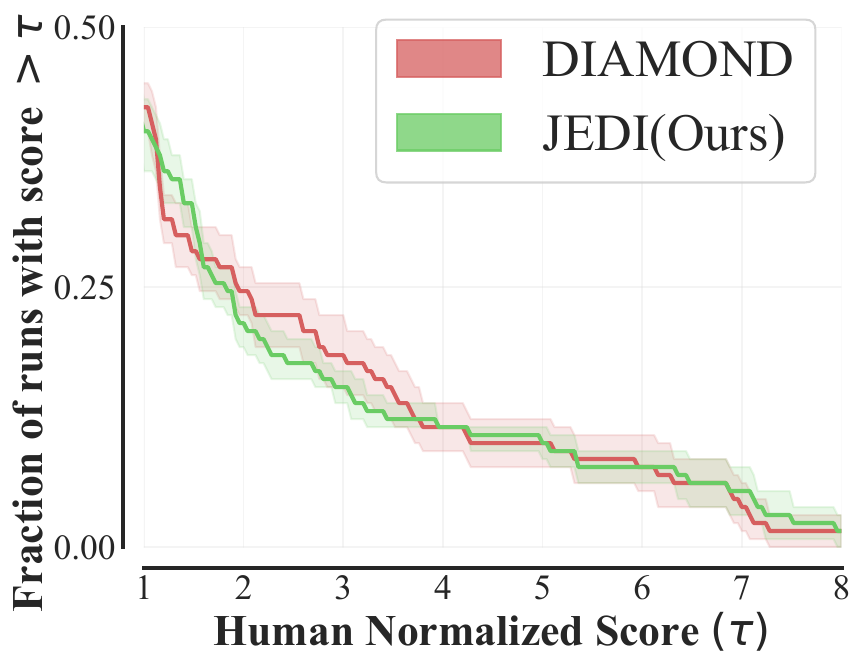}
  \end{subfigure}
  \caption{HNS performance profile. Left: low-HNS regime. Right: high-HNS regime. JEDI wins disproportionately often in low-HNS regimes, while DIAMOND's largest relative wins occur in higher-HNS regimes.}
  \label{fig:hns_performance_profile}
\end{figure}

\begin{table}[t]
  \centering
  \tiny
  \caption{Atari100k overall performance. Bold refers to SOTA. We report results for batch size 32 and 64. Green score means it is the higher score and we take that result. We retrieve the updated results of STORM from \cite{OC-STORM}. TWISTER does not publish seed-level results, so we cannot compute its exact IQM or optimality gap. JEDI achieves SOTA optimality gap and IQM and runner-up performance in number of super-human games. JEDI achieves SOTA results in 7 games.}
  \label{tab:atari100k_FULL_NUMBERS}

    \begin{tabular}{
        l                                
        rrrrrrrrrrr                   
    }
\toprule
Game                   &  Random    &  Human     &  TWM                &  IRIS      &  DreamerV3   &  STORM              &  DIAMOND            &  TWISTER            &  JEDI32/JEDI64                             &  JEDI(Ours)         \\
\midrule
Alien                  &  227.8     &  7127.7    &  674.6              &  420.0     &  1078.9      &  748.2              &  744.1              &  970.0              &  \textcolor{pastelgreen}{1090.1}/801.6     &  \textbf{1090.1}    \\
Amidar                 &  5.8       &  1719.5    &  121.8              &  143.0     &  141.2       &  144.0              &  \textbf{225.8}     &  184.0              &  114.3/\textcolor{pastelgreen}{120.2}      &  120.2              \\
Assault                &  222.4     &  742.0     &  682.6              &  1524.4    &  669.7       &  1376.7             &  \textbf{1526.4}    &  721.0              &  \textcolor{pastelgreen}{1394.5}/1388.3    &  1394.5             \\
Asterix                &  210.0     &  8503.3    &  1116.6             &  853.6     &  946.4       &  1318.5             &  \textbf{3698.5}    &  1306.0             &  2331.0/\textcolor{pastelgreen}{2626.3}    &  2626.3             \\
BankHeist              &  14.2      &  753.1     &  466.7              &  53.1      &  622.7       &  \textbf{990.0}     &  19.7               &  942.0              &  \textcolor{pastelgreen}{926.1}/653.3      &  926.1              \\
BattleZone             &  2360.0    &  37187.5   &  5068.0             &  13074.0   &  12400.0     &  5830.0             &  4702.0             &  9920.0             &  \textcolor{pastelgreen}{16084.0}/9056.0   &  \textbf{16084.0}   \\
Boxing                 &  0.1       &  12.1      &  77.5               &  70.1      &  73.8        &  81.2               &  86.9               &  88.0               &  \textcolor{pastelgreen}{91.6}/86.8        &  \textbf{91.6}      \\
Breakout               &  1.7       &  30.5      &  20.0               &  83.7      &  27.9        &  41.0               &  132.5              &  35.0               &  \textcolor{pastelgreen}{155.6}/128.3      &  \textbf{155.6}     \\
ChopperCommand         &  811.0     &  7387.8    &  \textbf{1697.4}    &  1565.0    &  411.7       &  1644.0             &  1369.8             &  910.0              &  1392.4/\textcolor{pastelgreen}{1589.2}    &  1589.2             \\
CrazyClimber           &  10780.5   &  35829.4   &  71820.4            &  59324.2   &  84880.0     &  79196.0            &  \textbf{99167.8}   &  81880.0            &  64786.2/\textcolor{pastelgreen}{69585.8}  &  69585.8            \\
DemonAttack            &  152.1     &  1971.0    &  350.2              &  2034.4    &  433.7       &  324.6              &  288.1              &  289.0              &  \textcolor{pastelgreen}{2047.2}/1933.6    &  \textbf{2047.2}    \\
Freeway                &  0.0       &  29.6      &  24.3               &  31.1      &  0.0         &  0.0                &  \textbf{33.3}      &  32.0               &  \textcolor{pastelgreen}{6.3}/0.0          &  6.3                \\
Frostbite              &  65.2      &  4334.7    &  \textbf{1475.6}    &  259.1     &  945.1       &  365.9              &  274.1              &  305.0              &  263.3/\textcolor{pastelgreen}{263.3}      &  263.3              \\
Gopher                 &  257.6     &  2412.5    &  1674.8             &  2236.1    &  5754.3      &  5307.2             &  5897.9             &  \textbf{22234.0}   &  \textcolor{pastelgreen}{4155.3}/4010.6    &  4155.3             \\
Hero                   &  1027.0    &  30826.4   &  7254.0             &  7037.4    &  11145.2     &  \textbf{11434.1}   &  5621.8             &  8773.0             &  \textcolor{pastelgreen}{7442.9}/6861.1    &  7442.9             \\
Jamesbond              &  29.0      &  302.8     &  362.4              &  462.7     &  480.4       &  408.0              &  427.4              &  \textbf{573.0}     &  \textcolor{pastelgreen}{460.5}/448.2      &  460.5              \\
Kangaroo               &  52.0      &  3035.0    &  1240.0             &  838.2     &  3790.0      &  3512.0             &  5382.2             &  \textbf{6016.0}    &  736.0/\textcolor{pastelgreen}{962.8}      &  962.8              \\
Krull                  &  1598.0    &  2665.5    &  6349.2             &  6616.4    &  7921.5      &  6522.2             &  8610.1             &  \textbf{8839.0}    &  8499.4/\textcolor{pastelgreen}{8676.9}    &  8676.9             \\
KungFuMaster           &  258.5     &  22736.3   &  \textbf{24554.6}   &  21759.8   &  24210.0     &  20046.0            &  18713.6            &  23442.0            &  12725.2/\textcolor{pastelgreen}{21340.6}  &  21340.6            \\
MsPacman               &  307.3     &  6951.6    &  1588.4             &  999.1     &  1358.4      &  1489.5             &  1958.2             &  \textbf{2206.0}    &  \textcolor{pastelgreen}{1075.9}/1024.4    &  1075.9             \\
Pong                   &  -20.7     &  14.6      &  18.8               &  14.6      &  18.9        &  18.4               &  \textbf{20.4}      &  20.0               &  18.2/\textcolor{pastelgreen}{18.4}        &  18.4               \\
PrivateEye             &  24.9      &  69571.3   &  86.6               &  100.0     &  1081.4      &  100.0              &  114.3              &  \textbf{1608.0}    &  \textcolor{pastelgreen}{95.2}/90.5        &  95.2               \\
Qbert                  &  163.9     &  13455.0   &  3330.8             &  745.7     &  3291.3      &  2910.5             &  \textbf{4499.3}    &  3197.0             &  \textcolor{pastelgreen}{3606.1}/3447.1    &  3606.1             \\
RoadRunner             &  11.5      &  7845.0    &  9109.0             &  9614.6    &  14995.0     &  14841.0            &  20673.2            &  17832.0            &  14690.0/\textcolor{pastelgreen}{25736.6}  &  \textbf{25736.6}   \\
Seaquest               &  68.4      &  42054.7   &  774.4              &  661.3     &  610.0       &  557.4              &  551.2              &  532.0              &  \textcolor{pastelgreen}{1103.7}/989.6     &  \textbf{1103.7}    \\
UpNDown                &  533.4     &  11693.2   &  \textbf{15981.7}   &  3546.2    &  7981.7      &  6127.9             &  3856.3             &  7068.0             &  4191.1/\textcolor{pastelgreen}{4271.4}    &  4271.4             \\
\midrule
\#SOTA (↑)           &  0         &  N/A       &  4                  &  0         &  0           &  2                  &  7                  &  6                  &  -                                         &  7                 \\
\#Superhuman (↑)     &  0         &  N/A       &  8                  &  10        &  9           &  11                 &  11                 &  \textbf{12}        &  \textcolor{pastelgreen}{11}/9             &  11                 \\
Mean (↑)             &  0.000     &  1.000     &  0.956              &  1.046     &  1.134       &  1.142              &  1.459              &  \textbf{1.621}     &  \textcolor{pastelgreen}{1.361}/1.358      &  1.450              \\
IQM (↑)              &  0.000     &  1.000     &  0.459              &  0.501     &  0.504       &  0.557              &  0.641              &  -                  &  0.609/\textcolor{pastelgreen}{0.618}      &  0.688              \\
Optimality Gap (↓)   &  1.000     &  0.000     &  0.513              &  0.512     &  0.500       &  0.489              &  0.480              &  -                  &  0.480/\textcolor{pastelgreen}{0.488}      &  0.460              \\
      \bottomrule
    \end{tabular}
\end{table}
\newpage


\newpage
\clearpage
\pagebreak

\end{document}